\documentclass{article}

\usepackage{microtype}
\usepackage[dvipsnames]{xcolor}
\usepackage{subcaption}
\usepackage{booktabs} 
\usepackage{minted}
\usepackage{needspace}
\usepackage{upgreek}

\usepackage{hyperref}

\usepackage[preprint]{icml2026}

\usepackage{amsmath}
\usepackage{amssymb}
\usepackage{mathtools}
\usepackage{amsthm}

\usepackage{booktabs}
\usepackage{multirow}
\usepackage{caption}
\usepackage{tcolorbox}
\usepackage{tikz,pgfplots}

\setminted{fontsize=\footnotesize,style=autumn}

\usepackage[capitalize,noabbrev]{cleveref}

\theoremstyle{plain}

\theoremstyle{definition}

\theoremstyle{remark}

\newcommand{\R}{\mathbb{R}}

\newcommand{\E}{\mathbb{E}}
\DeclareMathOperator*{\argmin}{arg\,min}
\newcommand{\microseconds}{\,\mathrm{\upmu s}}
\newcommand{\gbs}{\,\mathrm{GB/s}}

\usepackage[textsize=tiny]{todonotes}

\icmltitlerunning{1-Bit Wonder: Improving QAT Performance in the Low-Bit Regime}

\begin{document}

\twocolumn[
  \icmltitle{1-Bit Wonder: Improving QAT Performance in the Low-Bit Regime through K-Means Quantization
    }



  \icmlsetsymbol{equal}{*}

  \begin{icmlauthorlist}
    \icmlauthor{Sohir Maskey}{equal,aa}
    \icmlauthor{Constantin Eichenberg}{equal,aa}
    \icmlauthor{Johannes Messner}{equal,aa}
    \icmlauthor{Douglas Orr}{equal,gc}
  \end{icmlauthorlist}

  \icmlaffiliation{aa}{Aleph Alpha Research}
  \icmlaffiliation{gc}{Graphcore Research}

  \icmlcorrespondingauthor{Sohir Maskey}{sohir.maskey@aleph-alpha-research.com}

  \icmlkeywords{Machine Learning, ICML}
  \vskip 0.2in

\begingroup
\captionsetup{type=figure,width=\textwidth}
\centering

\hbox to \columnwidth{\hfil
  \includegraphics[width=\columnwidth]{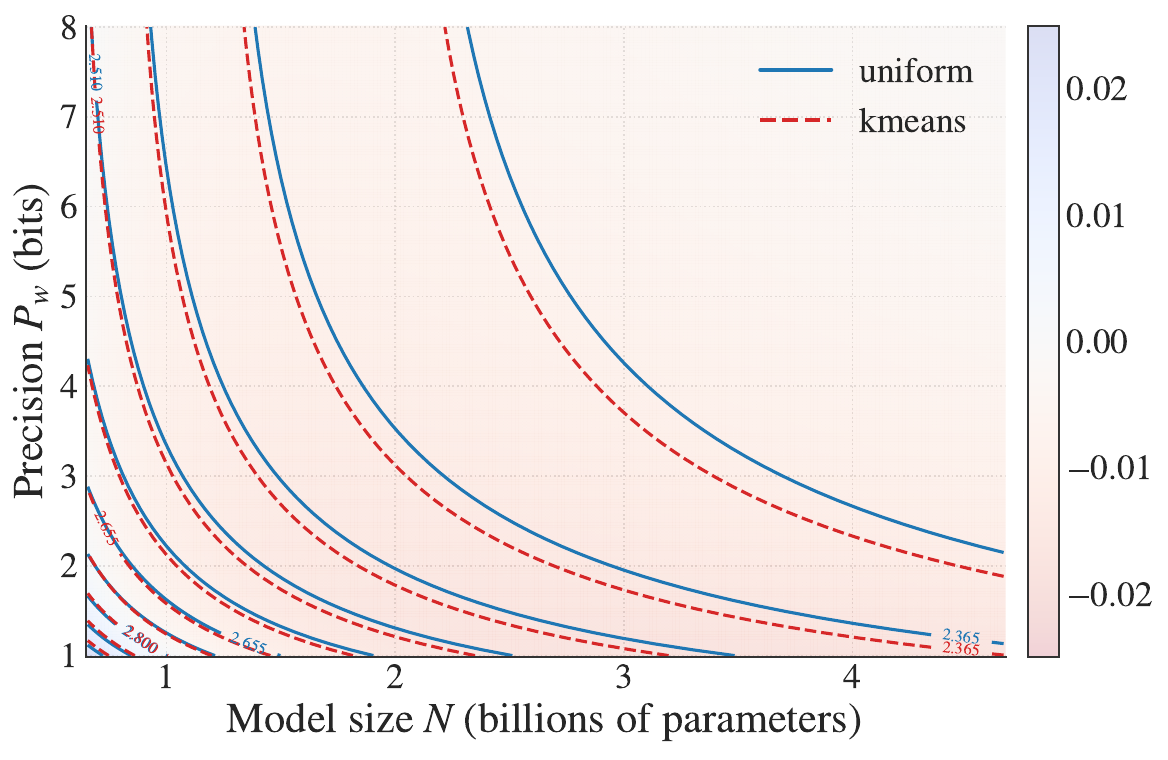}%
  \hspace{0.075\columnwidth}%
  
  \includegraphics[width=\columnwidth]{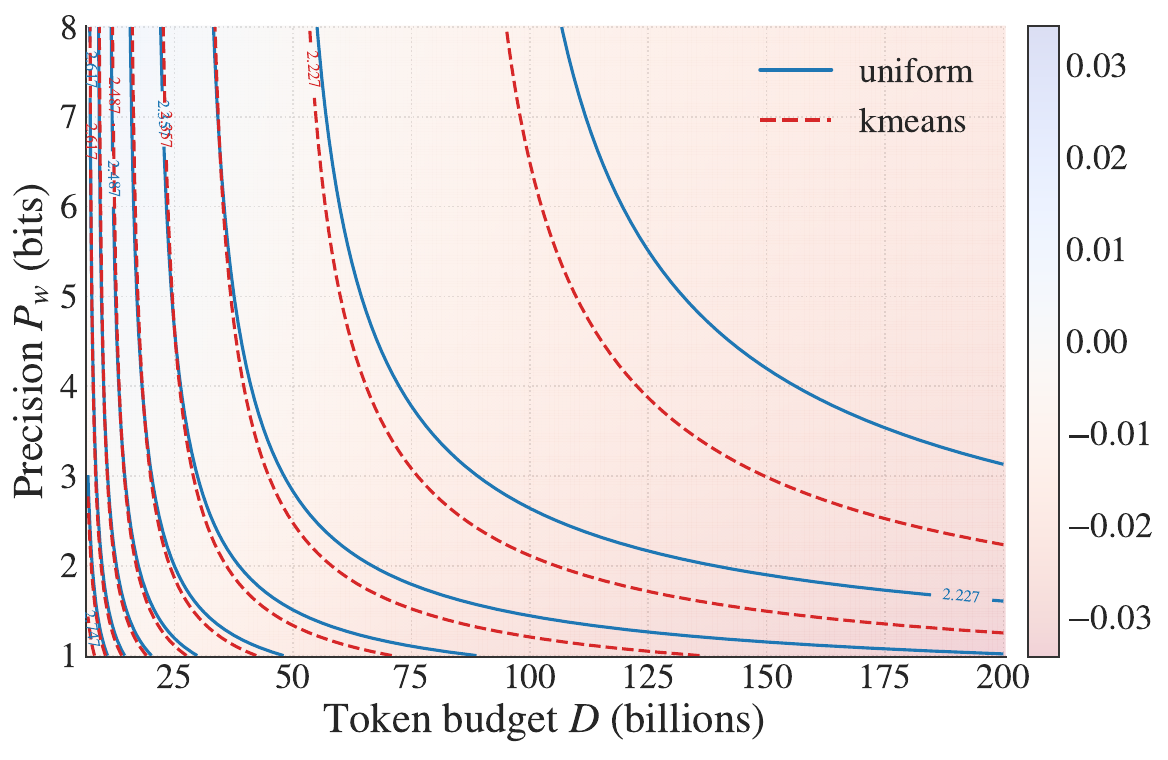}%
\hfil}\par

\captionof{figure}{\textbf{IsoLoss contours under precision--budget tradeoffs.}
(Left) Precision--parameter tradeoff under a fixed 50B token budget.
(Right) Precision--token tradeoff under a fixed 3.9B parameter budget.
Background colors show the predicted loss gap between uniform integer quantization and k-means quantization (red indicates lower loss for k-means).
Across all evaluated regimes, k-means strictly dominates uniform quantization, highlighting the consistent advantage of nonlinear formats under fixed memory budgets.}
\label{fig:isolosscurves}
\endgroup
\vskip 0.2in

]



\printAffiliationsAndNotice\,  

\begin{abstract}
  Quantization-aware training (QAT) is an effective method to drastically reduce the memory footprint of LLMs while keeping performance degradation at an acceptable level. However, the optimal choice of quantization format and bit-width presents a challenge in practice. The full design space of quantization is not fully explored in the context of QAT, and the precise trade-off between quantization and downstream performance is poorly understood, as comparisons often rely solely on perplexity-based evaluations. In this work, we address these shortcomings with an empirical study of QAT in the low-bit regime. We show that k-means based weight quantization outperforms integer formats and can be implemented efficiently on standard hardware. Furthermore, we find that, under a fixed inference memory budget, the best performance on generative downstream tasks is achieved with $1$-bit quantized weights.

\end{abstract}


\section{Introduction}
Large language models (LLMs) have demonstrated remarkable performance across a wide range of language tasks, largely driven by scaling model size \citep{brown2020gpt3, chowdhery2022palm, wei2022emergent}. However, this increase in parameter count comes at the cost of substantial memory footprint and bandwidth demands during both training and inference \citep{megatron-lm, Bommasani2021FoundationModels}. For moderate batch sizes, inference is typically memory-bound rather than compute-bound \citep{frantar2024marlinmixedprecisionautoregressiveparallel, mazurek2025llm}, making memory bandwidth and weight storage performance bottlenecks.

Recent work demonstrates that reducing weight precision can substantially alleviate inference bottlenecks in LLMs. For instance, \citet{frantar2024marlinmixedprecisionautoregressiveparallel} show that $4$-bit weight quantization enables near-ideal speedups for autoregressive LLM inference by mitigating memory bandwidth constraints. Motivated by such results, weight quantization has become a widely adopted approach for reducing inference-time costs, and is now employed in frontier LLMs such as Kimi-K2 \citep{kimi2025thinking}. By representing model weights in lower precision, quantization reduces both the memory footprint of the model and the volume of data transferred between memory and compute units, enabling practical speedups of memory-bound inference.

These memory reductions and speedups come with an inherent trade-off: quantization introduces approximation error that can adversely affect model performance. However, the extent to which accuracy degrades, particularly at very low bit-widths, remains an open and actively debated question. On the one hand, several works argue that quantization-aware training enables aggressive low-bit quantization with little to no loss in accuracy \citep{wang2023bitnetscaling1bittransformers, ma2024era1bitllmslarge}. On the other hand, empirical studies report that substantial performance degradation when reducing weight precision below 6 bits \citep{kumar2025scaling}. This discrepancy highlights the need for a more systematic and principled understanding of weight quantization.

To understand the source of these conflicting findings, a closer look at prior work reveals that much of the debate around aggressive low-bit quantization is driven by differences in evaluation methodology. Several influential works advocating ternary or $1$-bit weights primarily rely on training or validation loss, or token-level log-likelihood evaluations, to assess model quality \citep{wang2023bitnetscaling1bittransformers, ma2024era1bitllmslarge}. At the same time, studies reporting severe degradation at low bit-widths often restrict their analysis to similar loss-based metrics, without consistently incorporating state-of-the-art quantization techniques such as block scaling factors or nonlinear quantizers \citep{kumar2025scaling}. As a result, it remains unclear to what extent observed performance gaps reflect fundamental limitations of aggressive quantization, as opposed to artifacts of the evaluation protocol or quantization design choices. In particular, evaluations beyond loss are necessary, since recent work reports substantial degradation on more realistic downstream generative tasks even under moderate quantization \citep{li2024evaluatingquantizedlargelanguage}.

\paragraph{Contributions}
In this work, we revisit a central question underlying efficient LLM inference: \emph{given a fixed inference memory budget, how should one trade off parameter count and weight precision?} To this end, we contribute the following results:
\begin{enumerate}
    \item We conduct a precision-aware scaling law analysis 
    with two findings:
        \begin{enumerate}
            \item Using fewer bits is generally preferable under a fixed weight memory budget.
            \item A format based on scalar k-means quantization outperforms the commonly used symmetric integer format at low bit-widths.
        \end{enumerate}
    \item We complement our analysis with generative downstream evaluations and fair, memory-matched comparisons, pairing a $4$B parameter \texttt{bf16} model trained on $150$B tokens with its $4$-bit and $1$-bit counterparts of approximately $12$B and $31$B parameters, respectively.
    \item We demonstrate that efficient inference for the k-means format is feasible on standard hardware using vector lookup tables.
\end{enumerate}

Furthermore, we release our \href{https://github.com/Aleph-Alpha-Research/1-Bit-Wonder}{training code}, the \href{https://github.com/graphcore-research/fused-dequantisation-kernels}{kernels}, and the model checkpoints for our \href{https://huggingface.co/Aleph-Alpha-Research/1BW-Llama-4B-16Bit-SFT}{4B}, \href{https://huggingface.co/Aleph-Alpha-Research/1BW-Llama-12B-4Bit-SFT}{12B}, and \href{https://huggingface.co/Aleph-Alpha-Research/1BW-Llama-30B-1Bit-SFT}{31B} models, together with the corresponding inference code.

\section{Method}
This section presents our methodology, beginning with background on scalar quantization and quantization-aware training (QAT) in \cref{sec:background}, followed by a description of the specific quantization formats and QAT schedules used in our experiments in \cref{sec:quant_schemes}.

\subsection{Background on Quantization}
\label{sec:background}

\paragraph{Scalar quantization}
A \emph{quantization format} is specified by the components needed to quantize real-valued weight tensors. Concretely, for the scope of this work we define a format as a tuple $(\lambda, f)$: a \emph{scale function} $\lambda \colon \R^n \to \R_+$ and a \emph{quantizer} $f: \R \to \R$.  The quantizer $f$ is a piecewise constant function

\[
f(x) = \begin{cases} 
    q_0 & \text{if } x < x_0, \\
    q_i & \text{if } x \in [x_i, x_{i+1}), \\
    q_{N-1} & \text{if } x \geq x_N.
\end{cases}
\]

where $I =[x_0, x_N]$ defines an interval outside of which values get clipped to boundary bins $q_0$ or $q_{N-1}$, respectively. Because weight quantization formats are usually symmetric, we also require that $x_N = -x_0$.

Given a quantization format $(\lambda, f)$, the quantization and dequantization operations $Q$ and $\bar{Q}$ on a weight tensor $w \in \R^n$ are defined as 
\begin{align*}
    Q(w) = f( w /\lambda(w)), \quad
    \bar{Q}(w) = Q(w)\cdot \lambda(w),
\end{align*}
where $f$ is applied entrywise. In practice, $\lambda$ is some global statistic of $w$ chosen such that the majority of entries of $w$ get mapped to $I$, e.g., $\lambda \propto \mathrm{absmax}$.

The reconstructed tensor $\bar{Q}(w)$ is an approximation of the original tensor $w$. To materialize $\bar{Q}(w)$ we need to store both $Q(w)$ and $\lambda(w)$. Since entries of $Q(w)$ are by definition contained in the set $\{q_0,...,q_{N-1} \}$, we can store $Q(w)$ in $n = \log_2 N$ bits per entry, which is called the \emph{bit-width} of the format.

\paragraph{Rounding to nearest centroids} 
\label{par:rounding}
A special case of the quantizer $f$ is given by
\begin{equation}
\label{eq:centroid_quantizer}
    f_c(x) = \argmin_{c_i} |x - c_i|,
\end{equation}
for some set of centroid points $c = \{c_i\} \subset  I$, with $i=0,...,N-1$. In other words, $f_c$ is the function that rounds the input $x$ to its nearest neighbor among the centroids. \citet{Panter1951QuantizationDI} show that for a given probability distribution $p$ on $I$ and fixed number of quantization levels $N$, the piecewise constant minimizer of the reconstruction error $\E_p[(x - f(x))^2]$ is indeed of the form $f_c$ for some $c$. In general there is no closed form expression for the optimal centroids, but they can be obtained algorithmically by 1D k-means clustering~\citep{lloyd1982quant}.

\paragraph{Block formats}Instead of applying quantize/dequantize operations to a whole tensor $w$ we can also split $w$ into disjoint blocks $w_J$ of $B$ elements each, and apply the operations separately on each $w_J$. This means we must also store all $\lambda(w_J)$ instead of only $\lambda(w)$, slightly increasing memory overhead at the benefit of generally more accurate quantization. If $n$ bits are used to store each $q_i$ and $m$ bits are used to store $\lambda(w_J)$, the average bit allocation $P_w$ for a block quantized tensor with $d$ elements and block size $B$ is
\begin{align} \label{eq:fractional_bit_width}
    P_w = \frac{1}{d} \left(d \cdot n + \frac{d}{B} \cdot m \right) = n + \frac{m}{B}.
\end{align}
In this sense, block wise quantization can be interpreted as using a fractional bit-width, e.g., a 4 bit format with block size 64 where each block scale is stored as a 16 bit number has an average bit-width of 4.25 bits per element. Hence we extend our notion of format to triples $(\lambda, f, B)$, with bit-width given by~\cref{eq:fractional_bit_width}.

\paragraph{Linear formats} The most commonly used quantization formats are linear (also called uniform) formats. These are formats with a centroid quantizer $f_c$ where $c$ is a uniformly spaced grid. This has the benefit that tensor contractions (typically matrix multiplications) between $\bar{Q}(w)$ and $\bar{Q}(v)$ can be reduced to operations on integer-like types, which are often natively supported by hardware. We describe concrete examples in \cref{sec:quant_schemes}.

\paragraph{Quantization-Aware Training} Quantization-aware training (QAT) simulates quantized weights (and, optionally, activations) during training by using $\bar{Q}(w)$ in the forward pass and the unquantized weights $w$ as a surrogate for gradient computation in the backward pass. Optimizing the training objective under quantization noise typically leads to improved performance compared to applying quantization only after training~\citep{pytorch2024quant}. In practice, simulated quantization is implemented by replacing
\begin{align}
    w \leftarrow w + \mathrm{SG}(\bar{Q}(w) - w),
\end{align}
where $\mathrm{SG}$ denotes the stop-gradient operation. Unlike with low-precision training, which promises training as well as inference speedups, the backward pass is executed in high precision. This means that QAT typically incurs additional compute and memory overhead compared to standard training. However, QAT is generally more stable than low-precision training and still yields speed and memory improvements at inference time, since $\bar{Q}(w)$ is sufficient to reproduce the forward pass and can be represented in fewer bits than $w$.

\subsection{Quantization Formats and QAT Schedule}
\label{sec:quant_schemes}

In this section, we describe the concrete quantization formats and training procedures used throughout our experiments.
All schemes are instances of the general quantization framework introduced in \cref{sec:background}, and differ only in the choice of quantizer $f$ and scale function $\lambda$. For all formats, we use the same block size $B=64$ and store $\lambda$ in 16 bits.

\paragraph{Uniform integer quantization}
Our baseline format is uniform integer quantization. The $n$-bit integer format uses a centroid quantizer $f_c$ as defined in~\cref{eq:centroid_quantizer} and a scale function $\lambda$ defined by:
\begin{align*}
    c &= \begin{cases}
        \{-(2^{n-1}-1), \dots, 2^{n-1}-1\} \subset \mathbb{Z}, \ n \geq 2, \\
        \{-1, 1\}, \ n = 1,
    \end{cases} \\
    \lambda(w) &= \begin{cases}
        \frac{\mathrm{absmax}(w)}{(2^{n-1}-1)}, \ n > 2, \\
        \mathrm{absmean}(w), \ n \leq 2,
    \end{cases}
\end{align*}
i.e., for formats of $\leq\!2$ bits, we switch from absmax to absmean scaling, otherwise most values would be rounded to zero, degrading performance. For the $1$-bit format, we observe severe optimization instabilities that frequently lead to \texttt{NaN}s early in training. To mitigate this, we follow the normalization strategy introduced in BitNet \citep{wang2023bitnetscaling1bittransformers} and subtract the tensor-wise mean before quantization.

Regarding the bit-width of integer formats, we note that the number of centroids for $n > 1$ is $2^n - 1$. Hence, according to our definition, the bit-width is $P_w = \log_2 (2^n - 1)$ which is approximated by $n$ for large $n$, but significantly differs from $n$ in the low bit regime. In particular, at $n=2$ we have $c = \{-1, 0, 1 \}$ which effectively gives ternary weights at a bit-width of $P_w \approx 1.58$. Whether to view the actual bit-width as $\log_2 (2^n - 1)$ or $n$ depends on the perspective. From a theoretical point of view the former is more natural, while from a practical viewpoint one typically uses $n$ bits to store the quantized representations, since the logic to unpack non-bit-aligned values can be slow. For our scaling laws we argue it is more accurate to use $P_w = \log_2 (2^n - 1)$, as this corresponds to the most compact representation possible. Following \cref{eq:fractional_bit_width}, the average bit-width should also account for storing $\lambda$, using 16 bits per block of 64 values, an overhead of 0.25 bits per weight. The average bit-width for our $n$-bit integer formats is therefore:


\begin{center}
\small
\setlength{\tabcolsep}{5.5pt}
\begin{tabular}{cccccccc}
\toprule
\textbf{$n$} 
& 2 & 3 & 4 & 5 & 6 & 7 & 8 \\
\midrule
$P_w$
& 1.83 & 3.06 & 4.16 & 5.22 & 6.23 & 7.24 & 8.24 \\
\bottomrule
\end{tabular}
\end{center}

For simplicity, we still refer to these formats as the $n$-bit uniform format, despite the subtleties previously discussed.

\paragraph{Nonlinear k-means quantization}
Our second quantization format is based on nonlinear centroid quantization using 1D k-means clustering. For each weight tensor, we compute a set of centroids
\[
c = \{c_0,\dots,c_{N-1}\} \subset [-1,1]
\]
by running k-means on the normalized weights, and define the quantizer as in~\cref{eq:centroid_quantizer}. This is motivated by the discussion in~\cref{par:rounding}, as this choice of $f_c$ minimizes the $L^2$ reconstruction error for a fixed number of quantization levels under mild assumptions~\citep{lloyd1982quant}.

As with uniform quantization, we use block-wise formats and store one scale parameter per block, again incurring an overhead of 0.25 bits per weight. The key difference lies in the choice of $c$: instead of a uniform grid, the centroids are learned and adapt to the empirical weight distribution\footnote{Storing the set of centroids also increases the average bit-width. However, since $N \ll \mathrm{numel}(w)$, this contribution is negligible.}. Importantly, we find that this nonlinear format remains stable even in the $1$-bit regime, without requiring mean-shifting or other stabilization tricks.

We refer to this format as the $n$-bit k-means format.

\paragraph{Quantization-aware training schedule}
For all experiments, we employ QAT as described in Section~\ref{sec:background}. Rather than enabling QAT from the very start of training, we first perform a warm-up phase of 1\,000 training steps using standard \texttt{bf16} mixed precision. This design choice is motivated by two considerations. First, early training is characterized by a highly non-stationary and ``chaotic'' phase in which weight distributions evolve rapidly. Applying quantization during this phase can lead to unstable optimization and, in the case of k-means quantization, to trivial centroid configurations that do not reflect meaningful weight structure. Second, recent work has shown that the timing of when QAT is introduced can have a substantial impact on final model performance \citep{liu2025paretoq}. To isolate the effect of the quantization format itself, we therefore keep the QAT start point fixed across all experiments.

After step 1\,000, quantization remains active for the rest of training. For k-means quantization, the centroids are learned at the onset of QAT and then frozen. This allows QAT to adapt to a fixed set of centroids, improving training stability. 

\section{Experimental Results}
In this section, we present our experimental results for QAT. 
In \cref{sec:scaling_laws}, we derive scaling laws that explicitly account for the precision at which models are trained, including an analysis in weight-memory-matched settings. 
In \cref{sec:memory_matched_downstream}, we address our main research question by evaluating memory-matched models: a $4$B \texttt{bf16} baseline and its $12$B $4$-bit and $31$B $1$-bit counterparts. 
These models are trained in longer runs, followed by an SFT phase, and evaluated on downstream generative benchmarks. Finally, \cref{sec:efficiency} provides benchmarks of specialized kernels designed for efficient inference with the k-means format.

All models are trained with backbone-only weight quantization; embeddings and output projections remain in \texttt{bf16}. Additional details on the training setup, data mixture, hardware, architecture, and further ablations are provided in \cref{app:experiments}.\footnote{We will release our full training code, together with the final checkpoints of the $12$B and $31$B models.}

\subsection{Scaling Laws}
\label{sec:scaling_laws}

\paragraph{Experimental setup}
We train and evaluate a suite of language models on Nemotron-CC \citep{su2025nemotroncctransformingcommoncrawl}, using a standard Llama-style dense Transformer architecture~\citep{grattafiori2024llama3herdmodels}. 

Our experimental grid spans model size, training data, precision, and quantization format. Specifically, we consider parameter counts
\[
N \in \{0.8, 1.4, 3.9\}\ \text{(billions)},
\]
training token budgets
\[
D \in \{8.4, 16.8, 25.2, 33.6, 41.9, 50.3\}\ \text{(billions)},
\]
and weight precisions
\[
P_w \in \{1.25, 2.25, 3.25, 4.25, 6.25, 8.25\}
\]
for the k-means format, and
\[
P_w \in \{1.25, 1.83, 3.06, 4.16, 6.23, 8.24\}
\]
for the uniform format. 

All models are trained from scratch following the QAT schedule described in~\cref{sec:quant_schemes}. Evaluation is performed using pretraining loss, which provides an unbiased estimate of test loss in the infinite-data regime (training tokens $\ll$ total corpus size).


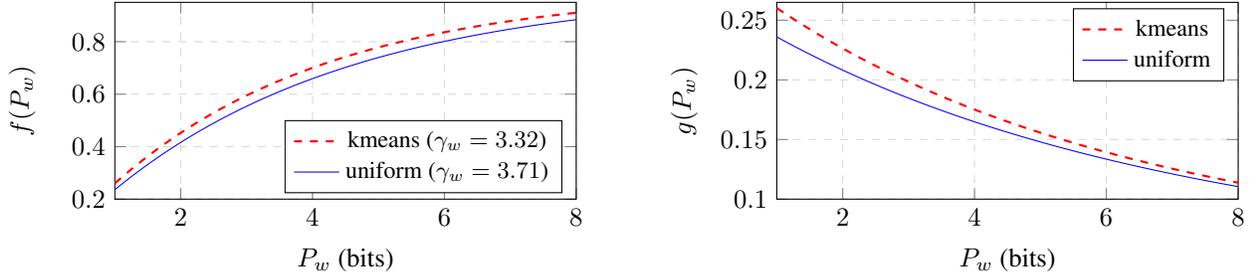
\begin{figure*}[t]
\centering
\begin{tikzpicture}

\begin{axis}[
    at={(-2cm,0)},
    anchor=north west,
    width=0.45\textwidth,
    height=4.2cm,
    xlabel={$P_w$ (bits)},
    ylabel={$f(P_w)$},
    xmin=1, xmax=8,
    ymin=0.2, ymax=0.95,
    grid=major,
    grid style={dashed, gray!30},
    legend pos=south east,
    legend style={font=\footnotesize},
]

\addplot[red, dashed, thick] table[x=Pw, y=f_Pw, col sep=comma]
    {tikz_export/f_bits_kmeans.csv};
\addlegendentry{kmeans ($\gamma_w=3.32$)}

\addplot[blue] table[x=Pw, y=f_Pw, col sep=comma]
    {tikz_export/f_bits_uniform.csv};
\addlegendentry{uniform ($\gamma_w=3.71$)}

\end{axis}

\begin{axis}[
    at={(6.8cm,0)},
    anchor=north west,
    width=0.45\textwidth,
    height=4.2cm,
    xlabel={$P_w$ (bits)},
    ylabel={$g(P_w)$},
    xmin=1, xmax=8,
    ymin=0.1, ymax=0.265,
    grid=major,
    grid style={dashed, gray!30},
    legend pos=north east,
    legend style={font=\footnotesize},
]

\addplot[red, thick, dashed] table[
    x=Pw,
    y expr=\thisrow{f_Pw}/\thisrow{Pw},
    col sep=comma
]{tikz_export/f_bits_kmeans.csv};
\addlegendentry{kmeans}

\addplot[blue] table[
    x=Pw,
    y expr=\thisrow{f_Pw}/\thisrow{Pw},
    col sep=comma
]{tikz_export/f_bits_uniform.csv};
\addlegendentry{uniform}

\end{axis}

\end{tikzpicture}

\caption{Precision-to-capacity mapping and the induced memory-normalized efficiency. 
(Left) The saturating function $f(P_w)$ models diminishing returns as weight precision increases, where higher values are better. 
(Right) The ratio $g(P_w)=f(P_w)/P_w$ determines the optimal precision under a fixed inference-memory budget $M = N P_w$, where higher values are better.}
\label{fig:f_and_g_bits_overlay}
\end{figure*}

\paragraph{Scaling law model with low precision parameters}
Following \citet{kaplan2020scalinglawsneurallanguage, hoffmann2022trainingcomputeoptimallargelanguage}, we adopt the standard decomposition of language model loss into three terms: one relating to model capacity, another to the amount of training data $D$ and a final irreducible loss term $E$. However, instead of treating the raw parameter count $N$ as the sole proxy for model capacity, we explicitly incorporate weight precision by introducing a \emph{effective parameter count} $N_{\mathrm{eff}}$, as suggested by \citet{kumar2025scaling}. Concretely, we model the loss as
\begin{equation}
\label{eq:final_scaling_mixed}
\mathcal{L}(N,D,P_w)
\;=\;
A\,N_{\mathrm{eff}}(N,P_w)^{-\alpha}
\;+\;
B\,D^{-\beta}
\;+\;
E,
\end{equation}
where $A,B,E,\alpha,\beta>0$ are fitted constants.

To capture the diminishing returns of increasing bit-width, we map the weight precision $P_w$ to effective capacity using a saturating function
\begin{equation}
\label{eq:saturation_f}
f(P_w)
\;=\;
1-\exp\!\left(-\frac{P_w}{\gamma_w}\right),
\end{equation}
with a learned scale parameter $\gamma_w>0$. The effective parameter count is then defined as $N_{\mathrm{eff}}(N,P_w) \;=\; N\,f(P_w)$.
This formulation ensures that reducing precision lowers the effective capacity of the model, while recovering the classical scaling law behavior as $P_w$ increases and $f(P_w)\to 1$.






\paragraph{Fit quality and parameter estimates}
All parameters are estimated via L-BFGS, see \cref{subsec:scaling_law_fitting_procedure} for more details. Both quantization formats are well described by the proposed scaling law, achieving $R^2 \approx 0.96$ and low RMSE. \Cref{fig:fit_quality} plots predicted versus observed loss. However, the fitted parameters reveal systematic differences between linear and nonlinear quantization.

For k-means quantization, we obtain
\[
\alpha \approx 0.63, \qquad
\beta \approx 0.40, \qquad
\gamma_w \approx 3.32,
\]
indicating a strong dependence of loss on effective model capacity (large $\alpha$) and a relatively fast saturation of capacity with increasing bit-width (small $\gamma_w$).

In contrast, uniform quantization yields
\[
\alpha \approx 0.55, \qquad
\beta \approx 0.46, \qquad
\gamma_w \approx 3.71,
\]
corresponding to weaker scaling with model size and a slower saturation in precision. We plot the predictions of these fits in \cref{fig:isolosscurves}. 

\begin{tcolorbox}[colback=gray!5,colframe=black,boxrule=0.5pt]
\textbf{Key takeaway.}
K-means quantization achieves uniformly lower loss than uniform integer quantization across all precision--budget tradeoffs, with the largest improvements in the ultra-low-bit regime.
\end{tcolorbox}

\paragraph{Optimal precision under a memory budget}
\label{par:optimal_precision}
A central motivation of this work is to understand how to allocate a \emph{fixed inference memory budget} between model size and weight precision. Assuming the dominant cost is weight storage, the memory budget can be approximated by
\begin{equation}
\label{eq:memory_budget}
M \;=\; N\,P_w,
\end{equation}
where $N$ is the parameter count and $P_w$ the bits per weight.



Under the constraint in \cref{eq:memory_budget}, we can rewrite $N=M/P_w$ and obtain $N_{\mathrm{eff}}(M,P_w)
=
\frac{M}{P_w}\,f(P_w)$.
Holding the training budget $D$ fixed, the only dependence on $P_w$ is through $N_{\mathrm{eff}}(M,P_w)$, and minimizing the loss in \cref{eq:final_scaling_mixed} is therefore equivalent to maximizing
\begin{equation}
\label{eq:g_definition}
g(P_w)
\;:=\;
\frac{f(P_w)}{P_w},
\end{equation}
which can be interpreted as \emph{effective capacity per memory bit}. This corresponds exactly to the quantity visualized in \cref{fig:f_and_g_bits_overlay} (right), while \cref{fig:f_and_g_bits_overlay} (left) illustrates the saturating behavior of $f(P_w)$.

Across both quantization formats (uniform and k-means), the ratio $g(P_w)=f(P_w)/P_w$ decreases as $P_w$ increases, indicating that under a fixed memory budget the scaling law favors allocating memory to \emph{more parameters} rather than \emph{more bits per parameter}. In other words, the predicted optimum is to push toward the lowest precision and reinvest the saved memory into scaling up $N$.

At the same time, the nonlinear k-means format yields consistently larger values of $g(P_w)$ in the low-bit regime, implying higher effective capacity per memory bit than uniform quantization. This advantage is most pronounced at $P_w \leq 4.25$, where nonlinear formats strictly dominate uniform ones under memory-matched comparisons. 

\begin{tcolorbox}[colback=gray!5,colframe=black,boxrule=0.5pt]
\textbf{Key takeaway.}
Under fixed memory, the best regime is the lowest stable precision balanced by scaling up parameters. K-means formats yield higher effective capacity per bit than uniform formats.
\end{tcolorbox}





\subsection{Memory-Matched Scaling and Downstream Generative Performance}
\label{sec:memory_matched_downstream}

To understand the trade-off between model scale and weight precision under a fixed memory budget, we conduct longer training runs and evaluate the resulting models on a broad suite of downstream generative benchmarks. Unlike loss- or perplexity-based analyses, our goal is to directly assess how quantization choices affect \emph{generative performance}.

\paragraph{Training setup and data mixture}
We train a $4$B-parameter model in \texttt{bf16} precision on $150$B tokens. Following recent best practices for strong open-weight LLMs, we adopt a curriculum inspired by \citet{bakouch2025smollm3}, in which general web data is progressively mixed with higher-quality sources such as code and math early in training. This data mixture is designed to improve reasoning and structured generation capabilities while preserving general language modeling performance. For general web and code data, we use filtered versions of Nemotron-CC \citep{su2025nemotroncctransformingcommoncrawl} and Starcoder \citep{lozhkov2024starcoder2stackv2}, respectively. For high-quality mathematical data, we incorporate a mixture of FineMath-3+ and FineMath-4+ \citep{allal2025smollm2smolgoesbig}.

To enable chat-style and instruction-following behavior, we further perform a short supervised fine-tuning (SFT) phase using the Tulu 3 SFT Mixture \citep{lambert2024tulu3}. 

\paragraph{Memory-matched model variants}
The $4$B \texttt{bf16} baseline requires approximately 7.8GB of memory for model weights at inference time. To study how this memory budget can be reallocated between parameter count and precision, we train two additional models whose total weight memory closely matches this budget: i) a $12$B-parameter model with $4$-bit weights, and ii) a $31$B-parameter model with $1$-bit weights. See \cref{sec:model} for exact architectures.
Both models use QAT with nonlinear k-means quantization format as described in \cref{sec:quant_schemes}. Aside from model size and weight precision, training recipes and data mixtures are kept equivalent.

\paragraph{Evaluation protocol}
We evaluate all models on a diverse set of benchmarks covering commonsense reasoning, multiple-choice question answering, code generation, instruction following, and mathematical reasoning. These include log prob evals (MMLU \citep{hendryckstest2021}, HellaSwag \citep{zellers2019hellaswag}, PIQA \citep{seo2018phrase}, ARC \citep{allenai:arc}), and further generative evaluations, i.e., HumanEval \citep{chen2021evaluating}, MBPP \citep{austin2021program}, GSM8K \citep{cobbe2021gsm8k}, MMLU-PRO \citep{wang2024mmlu}, 
and AidanBench \citep{mclaughlin2025aidanbench}. We follow standard evaluation protocols and report accuracy or pass@1 as appropriate, except for AidanBench, where we report a coherence score computed by GPT-4-mini as a judge. 

\paragraph{Results}
\Cref{tab:evals_generative_same_size} reports the downstream performance of the three memory-matched models. Both low-bit models outperform the $4$B \texttt{bf16} baseline across all benchmarks. 

The $31$B $1$-bit model achieves the strongest overall results on most knowledge, reasoning, and code-generation tasks, including MMLU, ARC-Challenge, MBPP, HumanEval, and MMLU-PRO. The $12$B $4$-bit model remains competitive and even performs best on GSM8K.

Overall, these results validate the scaling-law prediction from \cref{sec:scaling_laws}: under a fixed inference memory budget, allocating capacity to \emph{more parameters} rather than \emph{more bits per parameter} yields stronger downstream generative performance. They also demonstrate that learned nonlinear quantization formats enable even $1$-bit models to scale effectively without catastrophic degradation.

\begin{table}[t]
\caption{Memory-matched scaling evaluation results across model sizes and bit-widths.}
\label{tab:evals_generative_same_size}
\centering
\small
\begin{tabular}{lccc}
\toprule
{Benchmark} & {$4$B/$16$-bit} & {$12$B/$4$-bit} & {$31$B/$1$-bit} \\
\midrule
MMLU & $33.21$ & $50.86$ & $\mathbf{51.61}$ \\
HellaSwag & $50.16$ & $\mathbf{55.41}$ & $54.70$ \\
PIQA & $74.97$ & $76.88$ & $\mathbf{77.20}$ \\
ARC-C & $41.21$ & $49.15$ & $\mathbf{50.68}$ \\
ARC-E & $71.34$ & $75.25$ & $\mathbf{77.27}$ \\
\midrule
MBPP & $19.00$ & $26.40$ & $\mathbf{30.00}$ \\
HumanEval & $12.80$ & $15.24$ & $\mathbf{18.29}$ \\
HumanEval Instruct & $26.22$ & $35.98$ & $\mathbf{42.68}$ \\
GSM8K (5-shot) & $27.90$  & $\mathbf{48.52}$ & $45.26$ \\
IFEVAL & $51.16$ & $\mathbf{63.45}$ & $62.70$ \\
MMLU PRO COT & $9.86$ & $17.00$ & $\mathbf{17.71}$ \\
AidanBench & $88.08$ & $147.48$ & $\mathbf{167.83}$ \\
\bottomrule
\end{tabular}
\end{table}

\subsection{Efficiency}
\label{sec:efficiency}

Our model evaluations consider a fixed inference memory budget. This choice enables comparison across formats and models without reference to a particular hardware implementation or software stack. While model size during inference can be the limiting factor in some deployment scenarios, especially with the increased prevalence of mixture-of-expert (MoE) models, inference speed and energy consumption often dominate practical considerations. Accordingly, we evaluate the practical efficiency of nonlinear $4$-bit and ultra-low-precision $1$-bit weight formats with block scaling.

\paragraph{Hardware formats}
We distinguish between quantization formats with native hardware support and those requiring software implementation using generic hardware capabilities. Formats with hardware support are constrained by the fundamental capabilities of the process and technology. However introducing new formats is expensive and the development cycle is long. We consider hardware formats to be out of scope for this work, and refer the reader to \citet{rouhani2023sharedmicroexponents,sharma2018bitfusion} for an evaluation of related formats. At a high level, we note that data movement is often a substantial contributor to energy use \citep{horowitz2014energyproblem}, such that our memory budget constraint can serve as a coarse proxy for energy cost.

\paragraph{Software formats (theoretical)}
We consider an indicative theoretical model for the speedup from software-supported formats that share a common computational format. In this model, the kernel execution time is the maximum of the compute and memory transfer times, each executed at a device-dependent peak rate. The resulting speedup under some mild assumptions (see \cref{sec:app-compute-model}) is
\begin{equation}
\label{eq:theoretical-speedup}
    \text{speedup}_{1 \rightarrow 2} = \frac{t_1}{t_2} = \frac{\max(1, P_{w1}\cdot \nu/(16\cdot m))}
        {\max(1, P_{w2}\cdot \nu/(16\cdot m))},
\end{equation}
where $t_1$ and $t_2$ denote execution times, $m$ is the batch size and $\nu \coloneqq \mathrm{R_{compute}}/\mathrm{R_{transfer}}$ is the device-specific ratio of compute rate (op/s) to memory transfer rate (bytes/s).

\begin{figure}
    \centering
    \includegraphics[width=\linewidth]{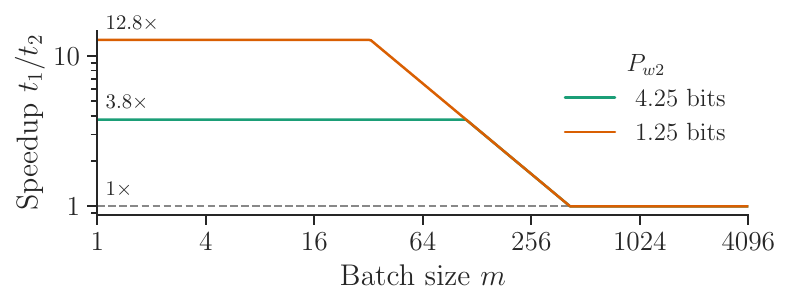}
    \caption{A theoretical model for speedup of $1$-bit and $4$-bit formats versus \texttt{bf16}, when decoded to \texttt{bf16} in software on an L40S GPU.}
    \label{fig:perf_theoretical}
\end{figure}

This relation highlights three regimes. For small $m$, the $\text{speedup}_{1\rightarrow 2}$ is the ratio of the precisions $P_{w1}/P_{w2}$, as both kernels are \emph{memory-bound} with execution time determined by the amount of data transferred from memory. For large $m$, the speedup is $1$ (no speedup), as both kernels become \emph{compute-bound} and perform approximately the same amount of arithmetic work. In the intermediate regime, the speedup scales $\propto 1/m$. For example, using \texttt{bf16} compute on the L40S GPU \citep{nvidia2024L40Sdatasheet}, this model predicts a peak speedup of $16/4.25 = 3.8\times$ for $4$-bit formats at batch size $m \leq 111$, with no speedup expected for $m > 418$. Similarly, $1$-bit formats achieve a peak speedup of $16/1.25 = 12.8\times$ when $m \leq 32$, again with no speedup when $m > 418$ (see \cref{fig:perf_theoretical}).

\paragraph{Software formats (practical)}
We evaluate the speed of fused dequantize-multiply operations on an inference-focused GPU, within micro-benchmarks and whole-model generation benchmarks (see \cref{sec:app-benchmarking} for full details). \Cref{tab:perf_microbenchmarks} shows micro-benchmark results for a setting representative of batch size $1$ autoregressive inference. Our custom Triton \citep{tillet2019triton} kernel uses a $256$-entry lookup table to support element formats of $\{1,2,4,8\}$ bits. It achieves near-theoretical-optimum speedups for $4$-bit nonlinear block-scaled formats, and a substantial speedup for $1$-bit formats relative to $16$ and $4$-bit baselines.

\begin{table}[t]
    \caption{Micro-benchmark results for the product of a $1 \times 8192$ \texttt{bf16} vector and $8192 \times 8192$ quantized matrix. The $4$-bit format achieves near-optimal speedup. Peak memory bandwidth for the test GPU is $864\gbs$. All standard errors are $<\!0.2\microseconds$.}
    \label{tab:perf_microbenchmarks}
    \centering
    \begin{tabular}{rrr}\toprule
        $P_w$ & Time (Speedup) & Effective BW \\\midrule
        $16$ & $175.6\microseconds$ ($1.0\times$) & $764\gbs$ \\
        $4.25$ & $49.5\microseconds$ ($3.7\times$) & $721\gbs$ \\
        $1.25$ & $24.0\microseconds$ ($7.6\times$) & $438\gbs$ \\
        \bottomrule
    \end{tabular}
    \vspace{-2mm}
\end{table}

Additional results for other dimensions, as well as a modified Marlin $4$-bit kernel \citep{frantar2024marlinmixedprecisionautoregressiveparallel}, which outperforms the Triton kernels for larger batch sizes, are provided in \cref{sec:app-benchmarking}. These show that, as expected, the advantage of weight quantization shrinks as batch size increases, with no speedup observed by batch size $256$, although the memory capacity benefit remains.

For whole-model token generation benchmarks, we adapt the Llama 3 model from Hugging Face by replacing all backbone linear layers with adapters that call the fused dequantize-multiply kernels evaluated above. We generate $100$ tokens from an empty KV cache and measure wall-clock time. To avoid measuring kernel launch overhead and host synchronization, we capture and replay the generation of a single token using a CUDAGraph, yielding high memory bandwidth utilization for the baseline.

Results are shown in \cref{tab:perf_modelbenchmarks}, demonstrating that the $12$B $4$-bit model from \cref{sec:memory_matched_downstream} runs only slightly slower than the $4$B $16$-bit baseline. The $31$B $1$-bit model is somewhat slower, due in part to the increased kernel overhead observed in \cref{tab:perf_microbenchmarks} and the extra compute from larger activations (e.g. RMSNorm and activation functions). While the larger $1$-bit models do not match $4$-bit throughput in this setting, they retain a substantial memory capacity advantage, enabling larger models under the same inference memory budget.

\begin{table}[t]
    \caption{Token generation performance for the models of \cref{sec:memory_matched_downstream} ({\color{blue}blue}), for single-user autoregressive generation. These correspond to effective bandwidth of $620$, $524$, and $372\gbs$ for $4$B $16$-bit, $12$B $4$-bit, and $31$B $1$-bit models respectively. The maximum error bar ($\pm2$ standard errors) is $<\!1\%$ of the mean.
    }
    \label{tab:perf_modelbenchmarks}
    \centering
    \begin{tabular}{rrrr}\toprule
        & \multicolumn{3}{c}{Tokens/s} \\
        $P_w$ & $4$B & $12$B & $31$B \\\midrule
        $16$ & {\color{blue}$88.8$} & $30.0$ & {\color{gray}OOM} \\
        $4.25$ & $190.1$ & {\color{blue}$79.3$} & $35.4$ \\\
        $1.25$ & $245.6$ & $120.0$ & {\color{blue}$60.7$} \\
        \bottomrule
    \end{tabular}
    \vspace{-5mm}
\end{table}

\section{Related Work}
\paragraph{Post-training quantization (PTQ)}
The most common approach to reduce the average weight precision of LLMs is PTQ, since it avoids expensive retraining of the full model.
However, simple uniform integer PTQ often degrades downstream performance, see, e.g., \citep{kim2024squeezellmdenseandsparsequantization}.
To mitigate this, a line of work develops \emph{sensitivity-aware} PTQ methods that explicitly account for which weights are most critical under quantization.

For example, OBQ \citep{frantar2022obc} casts layer-wise quantization as a quadratic error minimization problem and performs greedy weight quantization with error-compensating updates based on the inverse Hessian as sensitivity measure.
Building on this foundation, GPTQ-style methods \citep{frantar-gptq, frantar2023optq} scale these second-order ideas to LLMs by using efficient blockwise updates.

AWQ \citep{lin2023awq} estimates sensitivity through activation magnitudes and selectively preserves particularly important weights in \texttt{bf16}.
Going beyond uniform grids, SqueezeLLM \citep{kim2024squeezellmdenseandsparsequantization} combines Hessian-based sensitivity estimates with non-uniform quantization.
More broadly, recent work has emphasized that the {choice of quantization format itself} is a fundamental design axis: \citet{orr2025optimalformatsweightquantisation} systematically study optimal scalar quantization formats and show that nonlinear centroid-based representations can significantly improve reconstruction error compared to standard uniform grids.
Another complementary approach is SmoothQuant \citep{xiao2023smoothquant}, which enables joint weight and activation quantization to \texttt{int8} by smoothing activation outliers and migrating quantization difficulty from activations to weights via an equivalent transformation.

Overall, while modern PTQ methods can remain competitive down to $\sim$$4$-bit weights, their performance degrades rapidly below this regime, where quantization noise can no longer be absorbed without adapting model parameters. Moreover, even moderate quantization below $\sim$8 bits has been shown to substantially harm downstream generative abilities, particularly instruction following, self-calibration, and in-context learning \citep{li2024evaluatingquantizedlargelanguage}. 

\paragraph{QAT and ultra-low-bit models}
To enable aggressive quantization many approaches rely on QAT.
Representative examples include $1.58$-bit \citep{ma2024era1bitllmslarge} and $1$-bit weight networks \citep{wang2023bitnetscaling1bittransformers}, which demonstrate that, with appropriate training recipes (e.g., normalization), extremely low-bit weights can remain viable.
Despite this progress, frontier deployments have historically remained conservative (often $16$-bit), with a recent trend towards \texttt{int4} block formats \cite{kimi2025thinking}.

\paragraph{Scaling laws with precision constraints}
Scaling laws \citep{kaplan2020scalinglawsneurallanguage, hoffmann2022trainingcomputeoptimallargelanguage}   are a standard tool to extrapolate from low-resource experiments to regime-level decisions under fixed constraints (e.g., compute- or memory-budgets), typically modeling performance as a function of model size and data.
Most work treats parameter count as the primary proxy for capacity; an exception is \citet{kumar2025scaling}, which introduces precision-aware scaling law analyses and reports optima around $\sim$6 bits under their assumptions and evaluation protocol.
More recently, \citet{liu2025paretoq} study scaling with aggressively low precision down to $1$-bit, focusing primarily on linear quantization and additionally analyzing \emph{when} to introduce QAT during pretraining, finding that late-stage QAT can be advantageous.

\paragraph{Efficient inference with nonlinear formats}
Without native hardware support, practical speedups from low-bit nonlinear weight formats rely on custom kernels that fuse dequantization with matrix multiplication, keeping weights compressed in memory and expanding them only on chip. Early lookup table (LUT)-based methods such as SqueezeLLM \citep{kim2024squeezellmdenseandsparsequantization} demonstrated throughput gains from centroid-based quantization but fell short of ideal memory-bound scaling, with more recent methods such as FLUTE \citep{flute2024} employing offline weight restructuring and optimized LUT handling to approach bandwidth-limited performance for $4$-bit matmuls at small batch sizes.


\section{Discussion}

\paragraph{Limitations}
Our work advocates for nonlinear block-quantized and ultra-low-precision formats, based on a scaling laws analysis of popular formats. However, the space of formats is vast and the cost of scaling laws experiments prohibits a broad exploration. Importantly, there are many design decisions that might interact with this conclusion, such as the choice of learning rate and schedule during QAT. In terms of putting our findings into practice, our primary comparison in terms of memory capacity may be sufficient for some applications. However where the constraint is inference speed on current hardware, we observe speedups only for small batch sizes, which would not be efficient in many deployment scenarios, including those with substantial prefill compute.

\paragraph{Conclusion}
Our work demonstrates the strength of nonlinear formats, with substantial advantages over uniform quantization, especially below $4.25$ bits/parameter. We also demonstrate the effectiveness of ultra-low-precision formats using block scaling, with a $31$B parameter model with $1.25$ bits/parameter outperforming a $12$B parameter model with $4.25$ bits/parameter across a wide variety of tasks. Our hope is that these results will guide and inspire further work into ultra-low-precision formats and nonlinear format design.

\section*{Acknowledgments}

We thank Callum McLean and Luka Ribar for helpful discussions and support throughout the development of this work.

\section*{Impact Statement}

This paper presents work whose goal is to advance the field of machine learning. There are many potential societal consequences of our work, none of which we feel must be specifically highlighted here.

\bibliographystyle{icml2026}
\bibliography{example_paper}

\newpage

\appendix
\onecolumn

\section{Experimental Details}
\label{app:experiments}
All experiments were implemented in PyTorch~\citep{paszke2019pytorch}
using the \texttt{torchtitan} framework~\citep{liang2025torchtitan}.
All models are trained from scratch, following the QAT schedule described in the main paper. Our training code is publicly available at: \url{https://github.com/Aleph-Alpha-Research/1-Bit-Wonder}.

\subsection{Training Setup}
\label{app:training_setup}

\paragraph{Hardware}
All training runs were performed on $64$ NVIDIA H100 GPUs, each with $80$GB of DRAM.
We used Fully Sharded Data Parallelism (FSDP2).
For the $12$B and $31$B models, we additionally used activation checkpointing to reduce memory usage. 

\paragraph{Pretraining Configuration}
All experiments use a global batch size of $4\,194\,304$ tokens per optimization step, with a context length of $4\,096$.
We use the standard Llama~3 tokenizer with vocabulary size $128\,256$, and append \texttt{EOS} tokens at the end of documents.

\subsection{Pretraining Data}
\label{app:data_mix}
For all pretraining runs, we fix the data sampling seed to ensure that
models trained for the same number of steps deterministically see the
same tokens at each optimization step.

\paragraph{Scaling-law experiments}
All scaling-law runs in \cref{sec:scaling_laws} are trained on a filtered
subset of Nemotron-CC~\citep{su2025nemotroncctransformingcommoncrawl}.

\paragraph{Long-horizon pretraining curriculum}
For the long-horizon runs in \cref{sec:memory_matched_downstream}, we adopt
a curriculum inspired by \citet{bakouch2025smollm3}, in which general web
data is progressively mixed with higher-quality sources such as code and
mathematics. Concretely, we combine Nemotron with filtered
Starcoder-V2 data~\citep{lozhkov2024starcoder2stackv2} and FineMath-3+/4+
data~\citep{allal2025smollm2smolgoesbig}. The exact mixture schedule is
summarized in \Cref{tab:curriculum_mix}.

\begin{table}[t]
\caption{Pretraining curriculum mixture over data sources. Training runs
for $36$K steps with $4\,194\,304$ tokens per step ($\approx 151$B tokens total).
Phase lengths follow a $72\%/18\%/10\%$ split of total training. Each entry
reports both mixture proportion and the corresponding token count.}
\label{tab:curriculum_mix}
\centering
\small
\setlength{\tabcolsep}{4.5pt}
\begin{tabular}{lcc cc cc cc}
\toprule
& \multicolumn{2}{c}{{Nemotron}} 
& \multicolumn{2}{c}{{FineMath-3+}} 
& \multicolumn{2}{c}{{FineMath-4+}} 
& \multicolumn{2}{c}{{Starcoder-V2}} \\
\cmidrule(lr){2-3} \cmidrule(lr){4-5} \cmidrule(lr){6-7} \cmidrule(lr){8-9}
{Phase} 
& \% & Tokens 
& \% & Tokens 
& \% & Tokens 
& \% & Tokens \\
\midrule
Phase 1 (72\%) 
& 85\% & 92.4B 
& 3\%  & 3.3B 
& 0\%  & 0.0B 
& 12\% & 13.0B \\

Phase 2 (18\%) 
& 75\% & 20.4B 
& 0\%  & 0.0B 
& 16\% & 4.4B 
& 9\%  & 2.4B \\

Phase 3 (10\%) 
& 63\% & 9.5B 
& 0\%  & 0.0B 
& 27.4\% & 4.1B 
& 9.6\% & 1.4B \\

\bottomrule
\end{tabular}
\end{table}

\subsection{Model architectures}
\label{sec:model}
All models in this paper follow the Llama 3 architecture with hyperparameters scaled to match the target parameter count. Concretely, we consider a standard decoder-only Transformer \citep{vaswani2017attention} trained autoregressively with causal self-attention. The model consists of a stack of Transformer blocks, each comprising (i) pre-normalization using RMSNorm, (ii) multi-head masked self-attention with RoPE \citep{su2021roformer} applied to queries and keys, and (iii) a position-wise feed-forward network using a SwiGLU gating nonlinearity \citep{shazeer2020gluvariantsimprovetransformer}. Following the Llama 3 design, the attention module employs grouped query attention \citep{2305.13245}, i.e., a larger number of query heads than key/value heads, to improve computational efficiency while maintaining modeling capacity. Residual connections are used around both the attention and feed-forward sublayers.

Consistent with the Llama 3 design, we use no dropout and no bias terms in linear projections. 
Training is performed with a standard {causal language modeling mask}, ensuring that each token can only attend to past context. In addition, we apply {document-level masking} to prevent attention across document boundaries when processing concatenated sequences. Our implementation follows the modular reference provided in the \texttt{torchtitan} codebase for Llama 3 models.\footnote{\url{https://github.com/pytorch/torchtitan/tree/main/torchtitan/models/llama3}} 
Finally, we implement scaled dot-product attention (SDPA) using PyTorch's {FlexAttention} operator for efficient fused attention computation.\footnote{\url{https://github.com/meta-pytorch/attention-gym}}

\paragraph{Model scales}
\label{par:model_scales}
We scale model depth and width from the Llama~3 8B configuration to satisfy
a fixed inference weight-memory budget.
Our reference point is a $4$B-parameter \texttt{bf16} model, corresponding to
approximately $7.8$GB of weight storage (more precisely,
$3\,883\,551\,744$ parameters $\times$ 2 bytes; see the third row of
\Cref{tab:hparams} for the exact architecture).

For the memory-matched low-bit variants, we quantize only the Transformer
\emph{backbone} weights, while keeping the input embedding and output projection
in \texttt{bf16}.
For the $31$B $1.25$-bit model (see the last row of \Cref{tab:hparams}), we use
$d=6\,144$ and $48$ attention heads, satisfying standard divisibility constraints
($d$ is a multiple of $256$, $d$ is divisible by the number of heads, and the
number of heads is a multiple of the number of KV heads, i.e., $8$).
With untied embeddings, the embedding and output projection each contain
$128\,256 \times 6\,144 \approx 0.79$B parameters, for $1.58$B parameters in
\texttt{bf16} ($\approx 3.15$GB).
The remaining backbone contains $29.07$B parameters, which at $1.25$ bits/weight
requires $\approx 4.54$GB.
In total, this yields $\approx 7.7$GB, matching approximately the reference memory budget.
Similarly, the $12$B 4.25-bit model uses $d=4096$ with $L=48$ layers, for a total of
$11.5$B parameters.
Here, the backbone contains $10.47$B parameters, occupying $\approx 5.56$GB at
4.25 bits/weight, while the untied embedding and output projection contribute
$2 \times (128\,256 \times 4\,096) \approx 1.05$B parameters in \texttt{bf16}
($\approx 2.10$GB).
This again matches approximately the reference memory budget.

Similarly, the $12$B 4.25-bit model uses $d=4096$ with $L=48$ layers, for a total of
$11.5$B parameters.
Here, the Transformer backbone contains $10.5$B parameters, occupying roughly
$5.6$GB at 4.25 bits, while the embedding and output layers contribute about
$1.0$B parameters ($\sim 2.1$GB in \texttt{bf16}).
This again matches the reference memory budget. See \Cref{tab:hparams} for a detailed overview.

All models use grouped query attention with a fixed number of key/value heads ($n_{\mathrm{kv}}=8$ for models $\geq 1$B), RoPE with $\theta=500\,000$, and SwiGLU feed-forward blocks with multiplier $1.3$, with intermediate dimensions rounded to multiples of $1024$. Architectural hyperparameters for all scales are summarized in

\subsection{Optimization Hyperparameters}
\label{sec:hparams}
Across all runs, we optimized models using AdamW \citep{kingma2015adam} with weight decay $0.1$, and otherwise standard AdamW hyperparameters ($\beta_1=0.9$, $\beta_2=0.95$, $\epsilon=10^{-8}$). 
We employed a learning rate schedule consisting of a warmup phase over the first $100$ optimization steps, followed by a constant learning rate with a linear decay over the final $10\%$ of training. 
We apply gradient clipping with a global norm of $1.0$ to stabilize optimization at larger scales.

We experiment with maximum learning rates in the range $\{8\cdot 10^{-4}, 6\cdot 10^{-4}, 4\cdot 10^{-4}, 3\cdot 10^{-4}, 2\cdot 10^{-4}\}$.

\begin{table}[t]
\caption{Architectural and optimization hyperparameters for all model scales. All models use SwiGLU with multiplier $1.3$, RMSNorm pre-normalization, grouped query attention, FlexAttention-based SDPA, no dropout or bias terms, and gradient clipping at norm $1.0$.}
\label{tab:hparams}
\centering
\begin{tabular}{lccccccc}
\toprule
Model & Parameters N & Dim $d$ & Layers $L$ & Heads & KV Heads & Max LR & RoPE $\theta$ \\
\midrule
$0.8$B & $818\,714\,11$2 & $1\, 536$ & $12$ & $12$ & $6$  & $8\cdot 10^{-4}$ & $500\,000$ \\
$1.4$B   & $1\,431\,373\,824$ & $2\, 048$ & $16$ & $16$ & $8$  & $6\cdot 10^{-4}$ & $500\,000$ \\
$3.9/4$B   & $3\,883\,551\,744$ & $3\, 072$ & $24$ & $24$ & $8$  & $4\cdot 10^{-4}$ & $500\,000$ \\
$12$B  & $11\,520\,053\,248$ & $4\,096$ & $48$ & $32$ & 8  & $3\cdot 10^{-4}$ & $500\,000$ \\
$31$B  & $30\,643\,279\,872$ & $6\, 144$ & $60$ & $48$ & $8$  & $2\cdot 10^{-4}$ & $500\,000$ \\
\bottomrule
\end{tabular}
\end{table}

\subsection{Supervised Fine-Tuning (SFT)}
\label{app:sft}
After pretraining, all memory-matched models from \cref{sec:memory_matched_downstream} (i.e., $4$B \texttt{bf16}, $12$B $4$-bit and $30$B 1-bit models) undergo a short supervised fine-tuning phase on the Tulu~3 SFT mixture~\citep{lambert2024tulu3}. We use the Llama~3 instruction chat template together with the Llama~3 tokenizer. We train for five epochs, using the same optimizer settings as in pretraining but reducing the learning rate by a factor of $10^{-1}$. Moreover, we extend the sequence length to $8,192$ tokens and apply standard document packing to accelerate training.

\subsection{Additional Ablations}
We ablate the choice of block-wise normalization used in our quantization operators, comparing $\mathrm{absmean}$ scaling (normalization by the mean absolute value within each block) against $\mathrm{absmax}$ scaling (normalization by the maximum absolute value). 
This design choice has a substantial impact on uniform linear quantization at very low bit-widths. 
In particular, for extremely aggressive quantization ($\leq 2$ bits), $\mathrm{absmean}$ yields significantly lower training loss, whereas for moderate precision ($\geq 3$ bits), $\mathrm{absmax}$ becomes preferable. 

\begin{figure}[t]
    \centering
    \includegraphics[width=0.48\linewidth]{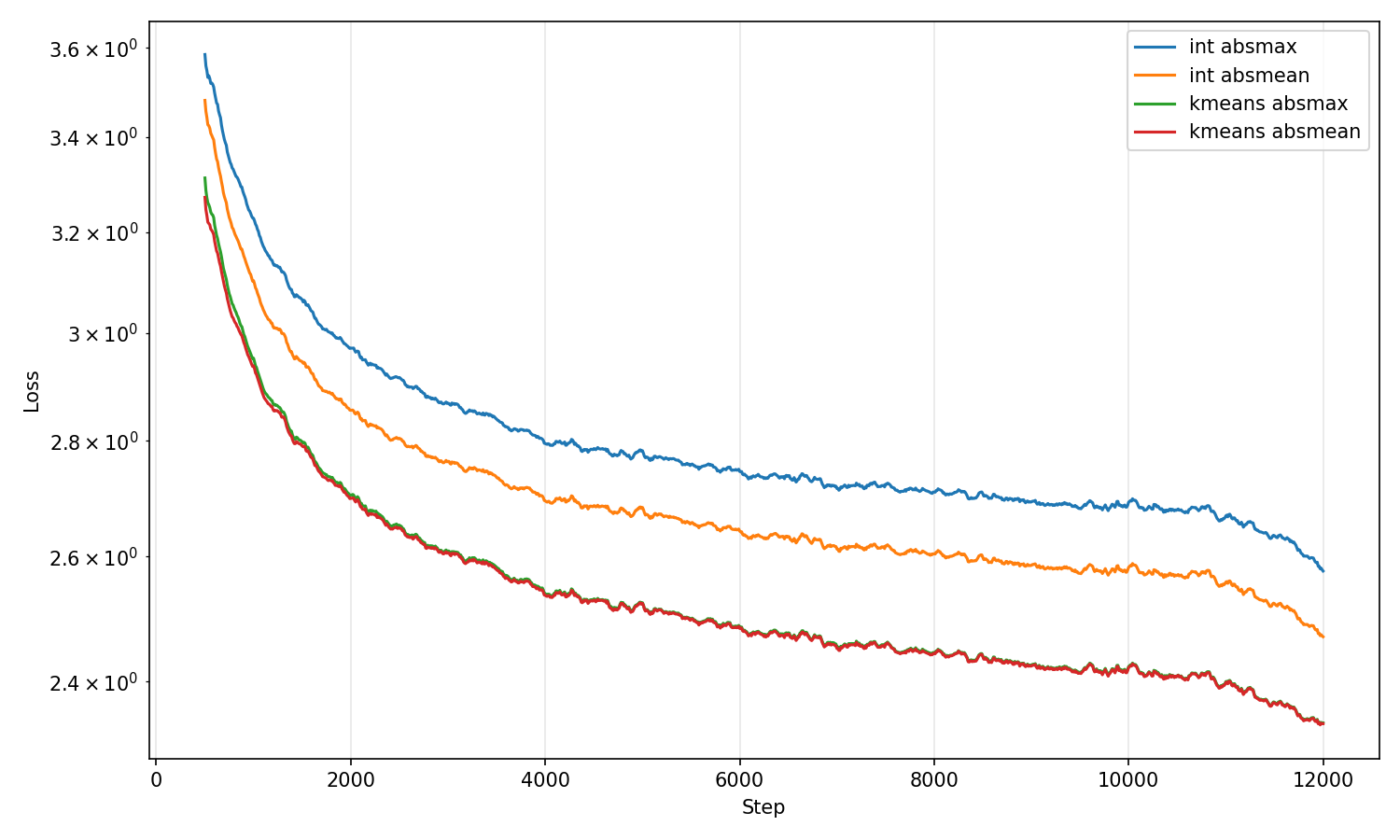}
    \hfill
    \includegraphics[width=0.48\linewidth]{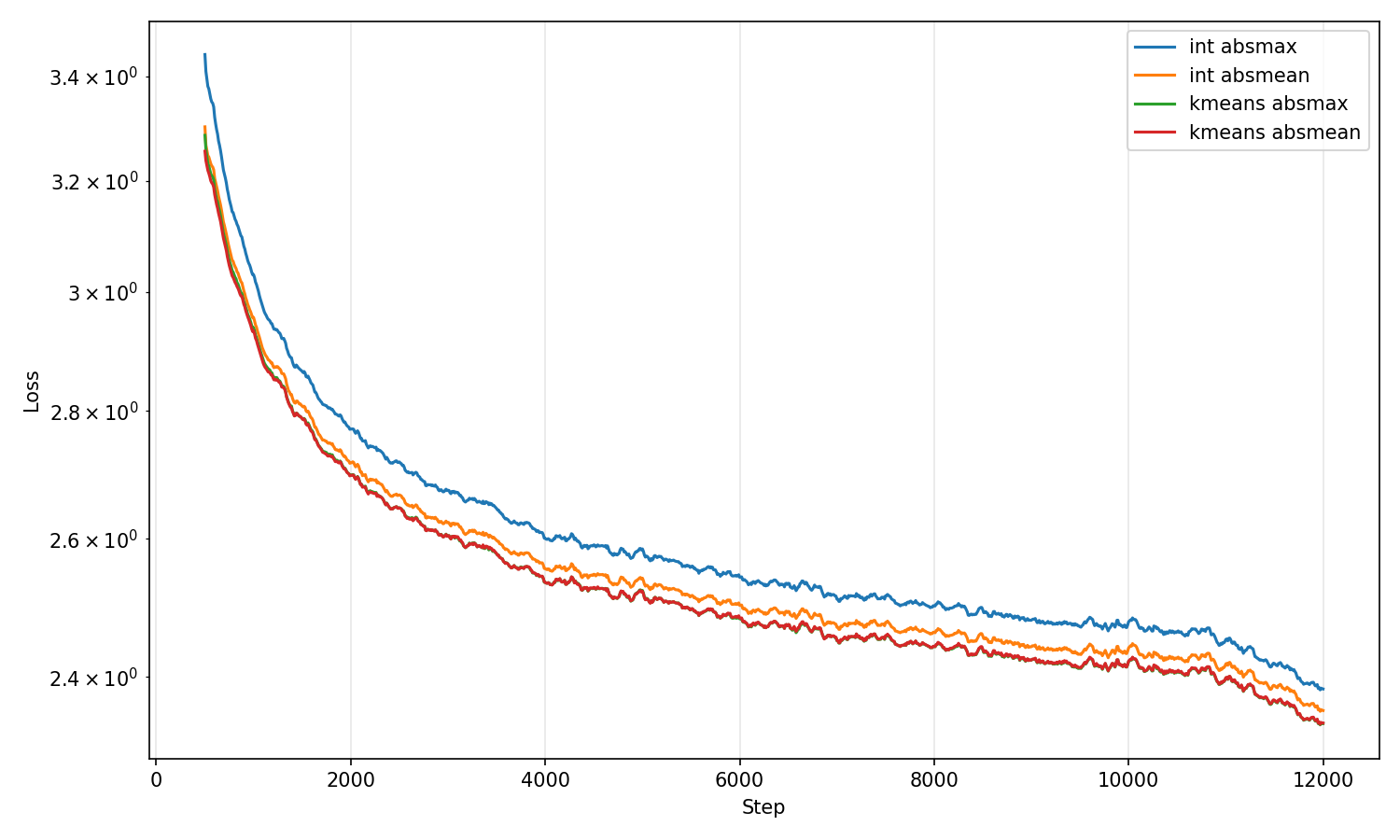}
    \caption{\textbf{Normalization ablation.} Training loss curves with logarithmic $y$-axis for different block-wise normalization strategies. 
    \textit{Left:} 2-bit quantization, where $\mathrm{absmean}$ scaling leads to substantially more stable optimization. 
    \textit{Right:} $4$-bit quantization, where $\mathrm{absmax}$ scaling becomes favorable. 
    K-means quantization remains largely insensitive to this choice.}
    \label{fig:norm_ablation}
\end{figure}

In contrast, the k-means quantizer is largely insensitive to the normalization choice, exhibiting similar training loss trajectories across all tested bit-widths.
\Cref{fig:norm_ablation} summarizes these findings, showing the training loss trajectories (log-scale) for 2-bit and $4$-bit quantization.

\subsection{Evolution of evaluation metrics}
\begin{figure}[t]
    \centering
    \includegraphics[width=0.48\linewidth]{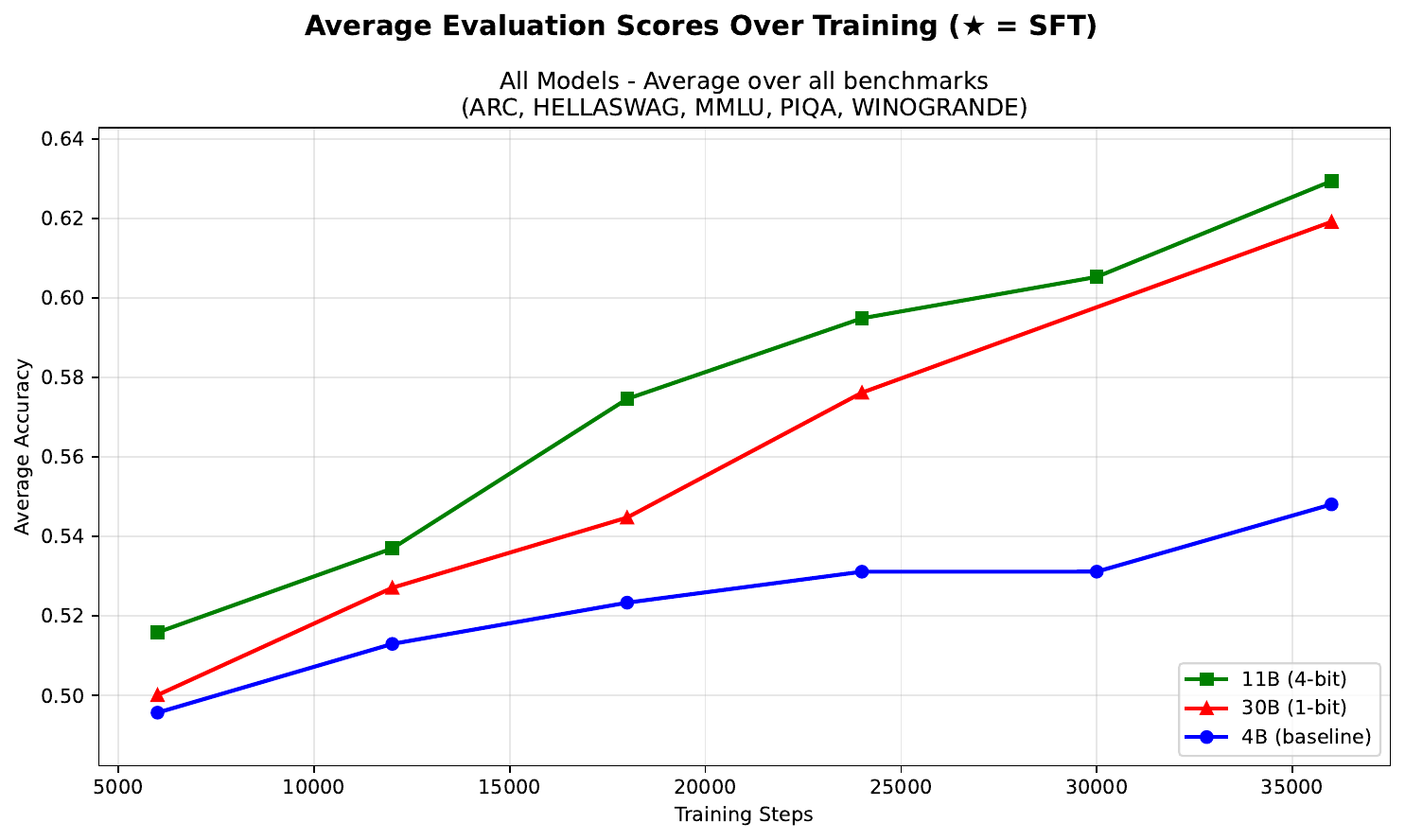}
    \hfill
    \includegraphics[width=0.48\linewidth]{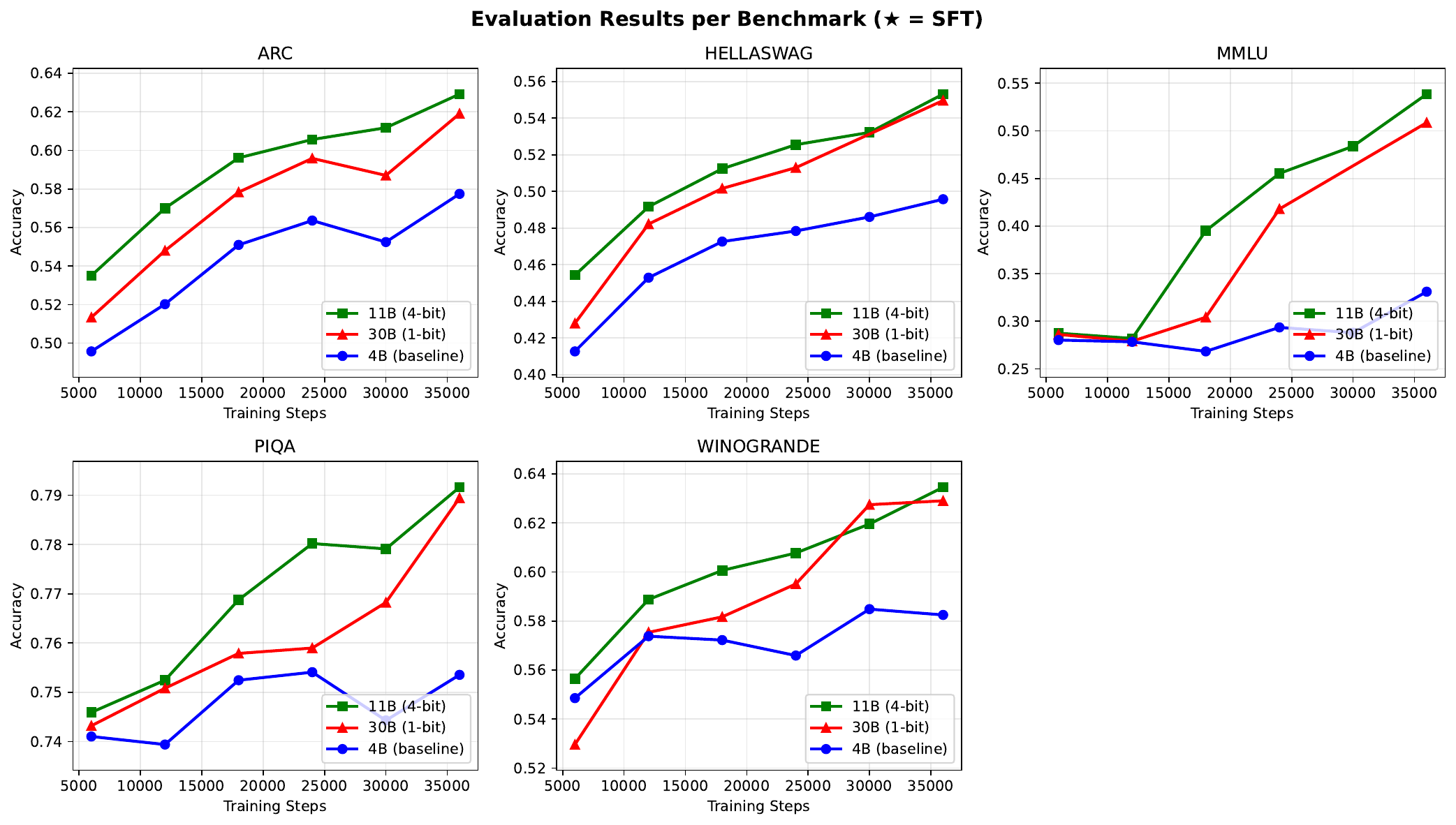}
    \caption{\textbf{Training progression} Evolution of pretraining evaluations during the training of the $4$B \texttt{bf16}, $12$B $4$-bit, and $31$B $1$-bit models.}
    \label{fig:eval_evolution}
\end{figure}

We perform logprob-based evaluations during the pretraining phase of our $4$B $16$-bit, $12$B $4$-bit, and $31$B $1$-bit models (\Cref{fig:eval_evolution}). We can see that here the $12$B 4bit model is consistently on-par with or better than the $31$B $1$-bit model. We therefore note that in our experiments these pretraining metrics are not an accurate predictor for downstream generative model performance, where the $31$B $1$-bit model is on-par with or better than the $12$B-$4$bit model (see \Cref{tab:evals_generative_same_size}).

\section{Scaling Laws}
Scaling laws characterize how the predictive performance of LLMs improves
as fundamental resources---most notably model size, data, and compute---are increased.
A central empirical finding is that, across wide regimes, test loss decreases
in a smooth and predictable manner, enabling reliable extrapolation and principled
trade-offs between scaling dimensions.
In particular, \citet{kaplan2020scalinglawsneurallanguage} and
\citet{hoffmann2022trainingcomputeoptimallargelanguage} report that loss follows
approximate power laws in model and dataset size.
A standard two-factor parametrization is
\begin{equation}
\label{eq:app:scaling_law}
\mathcal{L}(N,D)
\;=\;
A\,N^{-\alpha}
\;+\;
B\,D^{-\beta}
\;+\;
E,
\end{equation}
where $\mathcal{L}(N,D)$ denotes the (test) loss for a model with $N$ parameters
trained on $D$ tokens, and $A,B,E,\alpha,\beta>0$ are fitted constants.
The exponents $\alpha$ and $\beta$ quantify diminishing returns: increasing
$N$ or $D$ yields monotonically smaller, yet persistent, improvements.

\paragraph{Precision-aware scaling}
Recent work has extended \cref{eq:app:scaling_law} to account for quantization
effects by interpreting reduced numerical precision as a decrease in
\emph{effective} model capacity.
For example, \citet{kumar2025scaling} model QAT
by replacing $N$ with a precision-dependent effective size $N\,f(P_w)$,
where $P_w$ denotes the average weight precision.
In their formulation, $f(\cdot)$ is a saturating function in bits, such that
capacity approaches the full-precision regime as $P_w$ increases, i.e.,
$f(P_w)\approx 1$ for sufficiently large $P_w$.

\paragraph{Final precision-aware functional form}
In our setting, models are trained with weights quantized to an average
precision $P_w$ (including block-scale overhead).
We retain the standard decomposition into a model-size term and a data-size term,
but replace the raw parameter count by an effective parameter count
that depends on precision:
\begin{equation}
\label{eq:app:final_scaling_mixed}
\mathcal{L}(N,D,P_w)
\;=\;
A\,N_{\mathrm{eff}}(N,P_w)^{-\alpha}
\;+\;
B\,D^{-\beta}
\;+\;
E.
\end{equation}
To capture diminishing returns in bit-width, we map precision to effective
capacity via a saturating function
\begin{equation}
\label{eq:app:saturation_f}
f(P_w)
\;=\;
1-\exp\!\left(-\frac{P_w}{\gamma_w}\right),
\end{equation}
with fitted scale parameter $\gamma_w>0$.
The effective parameter count is then defined as
\begin{equation}
\label{eq:app:neff_simple}
N_{\mathrm{eff}}(N,P_w)
\;=\;
N\,f(P_w).
\end{equation}
This construction ensures that reducing precision lowers effective capacity,
while recovering the classical scaling law behavior as $P_w$ increases and
$f(P_w)\to 1$.

We empirically estimate $\mathcal{L}(N,D,P_w)$ using losses from models ranging
from $0.8$B to $3.9$B parameters, trained on datasets containing between $8$B and $50$B tokens.

\subsection{Fitting Procedure}
\label{subsec:scaling_law_fitting_procedure}
To estimate the parameters of the precision-aware scaling law in
\cref{eq:app:final_scaling_mixed}, we follow the robust parametric fitting
methodology of \citet{hoffmann2022trainingcomputeoptimallargelanguage}.
As in their work, we fit the scaling law in the log domain and optimize a robust
regression objective to mitigate the influence of outliers and heteroscedastic
noise in empirical loss measurements.

Specifically, we reparametrize the positive coefficients $A,B,E$ as
$a=\log A$, $b=\log B$, and $e=\log E$, and evaluate predicted log loss via a
log-sum-exp combination of the model and data terms:
\begin{equation}
\label{eq:log_law_pred}
\log \hat{\mathcal{L}}(N,D,P_w)
=
\log\!\sum\nolimits_{\exp}\!\Bigl(
a-\alpha\log N_{\mathrm{eff}}(N,P_w),\;
b-\beta\log D,\;
e
\Bigr),
\end{equation}
which ensures numerical stability and preserves the additive structure of the
original law.

Following \citet{hoffmann2022trainingcomputeoptimallargelanguage}, the fitting
objective is a Huber loss on the log-residuals,
\begin{equation}
\label{eq:huber_fit}
\min_{a,b,e,\alpha,\beta,\gamma_w}
\sum_{i}
\mathrm{Huber}_\delta\!\Bigl(
\log \hat{\mathcal{L}}(N_i,D_i,P_{w,i}) - \log \mathcal{L}_i
\Bigr),
\end{equation}
where $\mathcal{L}_i$ denotes the observed loss for the $i$-th run and
$\delta=10^{-3}$ is chosen to balance sensitivity and robustness.
The Huber loss blends squared and absolute residual penalties, making the fit
less sensitive to extreme observations while still maintaining high fidelity to
the bulk of the data.

We enforce positivity of $\alpha,\beta,\gamma_w$ via a softplus
reparameterization and optimize all parameters with the L-BFGS algorithm.
To mitigate sensitivity to initialization and local minima, we perform a
multi-start grid search over plausible initial values for
$(a,b,e,\alpha,\beta)$, selecting the solution with the lowest robust objective.

\begin{figure}
    \centering
    \includegraphics[width=\linewidth]{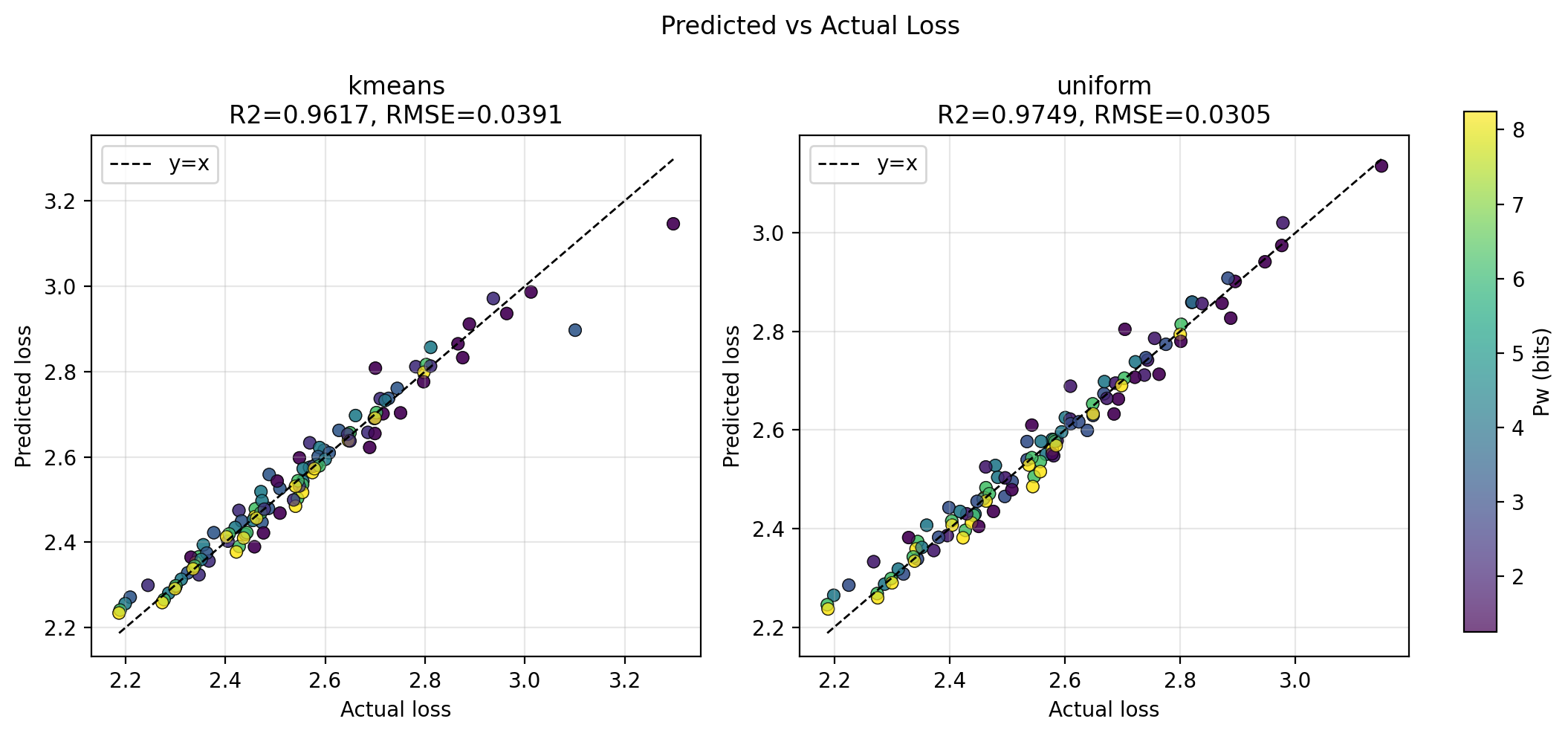}
    \caption{Predicted vs.\ actual loss for k-means (left) and uniform (right) quantization formats. Points are colored by precision $P_w$ (bits). Both methods achieve strong fits with $R^2 > 0.96$, with uniform quantization showing slightly better predictive accuracy (RMSE $= 0.0305$) compared to k-means (RMSE $= 0.0391$). The dashed line indicates perfect prediction ($y = x$).}
    \label{fig:fit_quality}
\end{figure}

\paragraph{Fit Quality}
Both quantization formats are well described by the proposed precision-aware
scaling law.
Across all runs, the fitted model achieves high predictive accuracy, with
$R^2 > 0.96$ and low RMSE.
\Cref{fig:fit_quality} compares predicted versus observed losses, showing that
the scaling law captures the systematic dependence on model size, token budget,
and weight precision.
While uniform quantization yields slightly lower RMSE, the overall fit quality
is comparable for both formats, supporting the use of the learned saturation
parameter $\gamma_w$ as a meaningful summary of precision-induced capacity loss.

\section{Extended Discussion on Scaling Laws}
\label{sec:neff-memory}

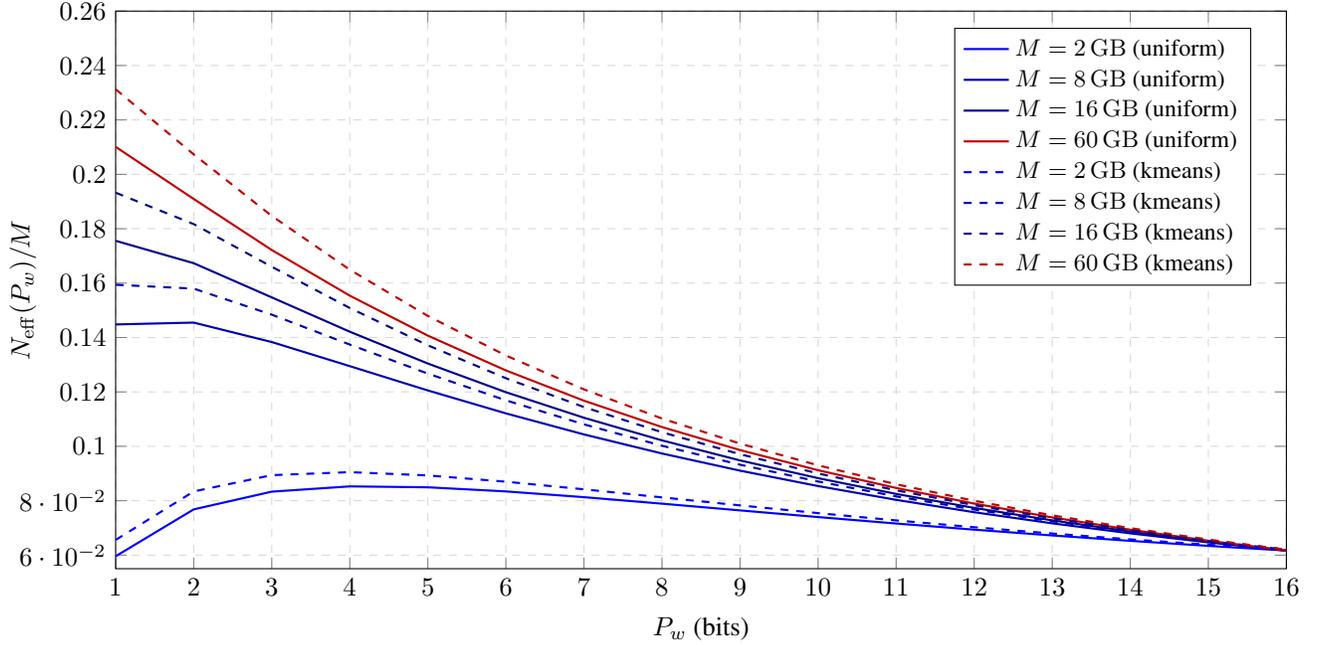
\begin{figure}[t]
\centering
\begin{tikzpicture}
\begin{axis}[
    width=\textwidth,
    height=9cm,
    xlabel={$P_w$ (bits)},
    ylabel={$N_{\mathrm{eff}}(P_w)/M$},
    xmin=1, xmax=16,
    ymin=0.055, ymax=0.26,
    grid=major,
    grid style={dashed, gray!30},
    legend pos=north east,
    legend style={font=\footnotesize, cells={anchor=west}},
]


\addplot[blue, thick] coordinates {
(1,0.059579473397490895) (2,0.07680776350986415) (3,0.08333025280561153) (4,0.08526744855180522)
(5,0.08493122854257872) (6,0.08341645828509082) (7,0.08130227945638634) (8,0.07891345881428473)
(9,0.07643724524634064) (10,0.07398209763291595) (11,0.07160963850006197) (12,0.06935303353410577)
(13,0.06722798742481427) (14,0.06523950634817662) (15,0.0633861349271705) (16,0.061662640597400876)
};
\addlegendentry{$M=2\,\mathrm{GB}$ (uniform)}

\addplot[blue!70!black, thick] coordinates {
(1,0.14480922231603405) (2,0.1454854786054133) (3,0.1383319189952825) (4,0.12946596116602607)
(5,0.12053534841003961) (6,0.11210563728195046) (7,0.10436693155237847) (8,0.09735726908764979)
(9,0.09104965280151044) (10,0.08539022542787382) (11,0.08031613153547547) (12,0.07576406943092578)
(13,0.07167429634179648) (14,0.06799232807934248) (15,0.06466945263093975) (16,0.061662640597400876)
};
\addlegendentry{$M=8\,\mathrm{GB}$ (uniform)}

\addplot[blue!55!black, thick] coordinates {
(1,0.17555570296249157) (2,0.16731959174391844) (3,0.15475159664151508) (4,0.1421290936590063)
(5,0.13043106839472696) (6,0.11989000724874495) (7,0.11050180646468935) (8,0.10218035332766147)
(9,0.09481448519867003) (10,0.08829095468757403) (11,0.08250377202969703) (12,0.0773576175799186)
(13,0.07276857088714912) (14,0.0686636327060894) (15,0.06497974345339168) (16,0.061662640597400876)
};
\addlegendentry{$M=16\,\mathrm{GB}$ (uniform)}

\addplot[red!75!black, thick] coordinates {
(1,0.21009399553187277) (2,0.1909324923447506) (3,0.17216890538499416) (4,0.15538885790642132)
(5,0.14069221039926683) (6,0.1278982884213314) (7,0.11677119011736649) (8,0.10708069138643281)
(9,0.09861997700549617) (10,0.09120941877729337) (11,0.084695461309463) (12,0.07894787011792093)
(13,0.07385660720496413) (14,0.06932883407164855) (15,0.06528622539315224) (16,0.061662640597400876)
};
\addlegendentry{$M=60\,\mathrm{GB}$ (uniform)}


\addplot[blue, thick, dashed] coordinates {
(1,0.06558182772769806) (2,0.08340476391110459) (3,0.08939631208918106) (4,0.09049793154876688)
(5,0.08929673481485671) (6,0.08698966466335951) (7,0.08418941399180549) (8,0.08122493810071395)
(9,0.07827522272168255) (10,0.07543584580025015) (11,0.07275463584667985) (12,0.07025175332526576)
(13,0.06793138547099073) (14,0.0657887037957448) (15,0.06381405030194312) (16,0.06199546120876646)
};
\addlegendentry{$M=2\,\mathrm{GB}$ (kmeans)}

\addplot[blue!70!black, thick, dashed] coordinates {
(1,0.15939807671599898) (2,0.1579811914979168) (3,0.14840184670080375) (4,0.13740767303925958)
(5,0.12673092368364397) (6,0.116907766099344) (7,0.10807311758375832) (8,0.10020899190224632)
(9,0.09323899401138204) (10,0.08706814329849112) (11,0.0816003407484385) (12,0.07674586741706448)
(13,0.07242421556352352) (14,0.06856469925634232) (15,0.06510603159400012) (16,0.06199546120876646)
};
\addlegendentry{$M=8\,\mathrm{GB}$ (kmeans)}

\addplot[blue!55!black, thick, dashed] coordinates {
(1,0.19324212202227886) (2,0.1816906313814449) (3,0.16601680138827482) (4,0.1508475884701339)
(5,0.1371352884672253) (6,0.12502558537563915) (7,0.11442584874005447) (8,0.10517335064074029)
(9,0.09709435396643722) (10,0.09002587188613882) (11,0.08382296036856692) (12,0.07836006575519373)
(13,0.07352993936694065) (14,0.06924165504153887) (15,0.06541841716814703) (16,0.06199546120876646)
};
\addlegendentry{$M=16\,\mathrm{GB}$ (kmeans)}

\addplot[red!75!black, thick, dashed] coordinates {
(1,0.23125998663451264) (2,0.20733163835616117) (3,0.18470201013014517) (4,0.16492073429067866)
(5,0.14792385813942333) (6,0.13337690726169077) (7,0.12091786519196505) (8,0.11021722606423424)
(9,0.10099135101002331) (10,0.09300168379320325) (11,0.08604969351201527) (12,0.0799709257757533)
(13,0.07462935967302357) (14,0.06991245618723944) (15,0.06572696814610128) (16,0.06199546120876646)
};
\addlegendentry{$M=60\,\mathrm{GB}$ (kmeans)}

\end{axis}
\end{tikzpicture}
\caption{Effective parameter density $N_{\mathrm{eff}}(P_w)/M$ versus backbone bit-width $P_w$ for fixed weight-memory budgets $M$ (GB). Solid lines: uniform quantization; dashed lines: kmeans quantization. The 16-bit special block consists of \emph{both} token embeddings and the LM head ($E(N)=2Vd(N)$). For large budgets the density is maximized at the smallest feasible $P_w$, while at very small budgets the optimum can shift upward because the 16-bit special block becomes non-negligible. Interestingly, the optimal bit-width for uniform formats at 8 GB memory-budget is 2 bits, while the kmeans format yields 1 bit as the optimal bit-width.}
\label{fig:neff_over_m_vs_p}
\end{figure}

This section provides an extended discussion of \Cref{par:optimal_precision}. While the main body of the paper focuses on the regime in which backbone parameters dominate the memory footprint, we now consider a more general setting. In particular, we present a comprehensive analysis of how to allocate a fixed inference memory budget between model size and weight precision.

\paragraph{Goal.}
We consider the common deployment setting where the dominant constraint is weight memory, i.e., only $M$ Gb (Gigabit) are available to store parameters. Lowering the backbone weight precision $P_w$ allows us to store more parameters, but extremely low precision yields diminishing returns. To capture this, we remember that we use the saturating quality factor
\begin{equation*}
    f(P_w)\;=\;1-\exp(-P_w/\gamma_w),
    \qquad \gamma_w>0,
\end{equation*}
and define the effective parameter count
\begin{equation*}
    N_{\mathrm{eff}}(P_w, N)\;=\; f(P_w)\,N,
\end{equation*}
where $N$ denotes the total number of parameters.

\paragraph{Memory accounting and the induced coupling $N(P_w)$.}
We store the embedding parameters in $16$ bits, while all remaining backbone parameters are stored in $P_w$ bits.
Let $E(N)$ be the number of embedding parameters for a model with total size $N$ (in billions).
Then the weight-memory budget in Gb is
\begin{equation}
    M \;=\; 16\,E(N) \;+\; P_w\bigl(N - E(N)\bigr)
    \;=\; P_w\,N + (16-P_w)\,E(N).
    \label{eq:mem_budget}
\end{equation}
For a fixed budget $M$ and a chosen precision $P_w$, the largest feasible parameter count $N$ is generally not free: it is implicitly defined by \eqref{eq:mem_budget} ($E(N)$ is generally unknown and there is generally no clear rule for estimating it based on $N$). We denote this maximal feasible size by $N(P_w)$.

\paragraph{Objective: maximize effective capacity per memory.}
Since $M$ is fixed, it is convenient to maximize effective capacity normalized by memory,
\begin{equation}
    \frac{N_{\mathrm{eff}}(P_w)}{M}
    \;=\;
    \frac{f(P_w)\,N(P_w)}{M},
    \qquad\text{and we seek}\qquad
    P_w^\star \in \arg\max_{P_w}\frac{f(P_w)\,N(P_w)}{M}.
    \label{eq:objective_neff_over_m}
\end{equation}
In general, there is no closed-form reduction because $E(N)$ depends on architecture and scales with $N$, so $N(P_w)$ must be obtained by solving \eqref{eq:mem_budget}. Practically, we evaluate a small set of supported precisions $\{P_w\}$; for each $P_w$ we solve \eqref{eq:mem_budget} for $N(P_w)$, then compute \eqref{eq:objective_neff_over_m}, and select the maximizer. We therefore require an explicit mapping $N\mapsto E(N)$ to evaluate \eqref{eq:mem_budget} and \eqref{eq:objective_neff_over_m}.

\paragraph{Hidden-size scaling from observed model classes.}
For Llama-style architectures, we fit a power law mapping total parameters $N$ to hidden size $d(N)$,
\begin{equation}
    d(N) \;\approx\; d_0\left(\frac{N}{N_0}\right)^{\alpha},
    \qquad
    N_0 = 3\,883\,551\,744,\;\; d_0=3072,\;\; \alpha \approx 0.320.
    \label{eq:d_of_N}
\end{equation}
The fit is anchored at the $3.88$B class and matches the following observed pairs $(N,d)$ in log-space:
\[
(3\,883\,551\,744,\;3072),\;
(11\,520\,053\,248,\;4096),\;
(30\,643\,279\,872,\;6144).
\]
(For implementation, $d$ is typically rounded to a hardware-friendly multiple; here we keep \eqref{eq:d_of_N} continuous.)

\paragraph{Embedding parameter count.}
With vocabulary size $V=128\,256$ and untied embedding and LM head matrices of shape $V\times d$, the embedding parameter count is
\begin{equation}
    E(N)
    \;=\;
    2\,V\,d(N)
    \;\approx\;
    2\cdot 128\,256 \cdot 3072 \cdot 10^{-9}\left(\frac{N}{3.88}\right)^{0.320}.
    \label{eq:E_of_N}
\end{equation}
Substituting \eqref{eq:E_of_N} into \eqref{eq:mem_budget} yields the implicit relation defining $N(P_w)$ for each $P_w$ and fixed $M$, and thus the effective density \eqref{eq:objective_neff_over_m}.

\subsection{Regime-dependent optimum}
\label{sec:regime-dependent}
We plot, for different memory budgets $M$ in $\mathrm{GB}$ (Gigabyte), the effective capacity per unit of memory as a function of the precision $P_w$ in \cref{fig:neff_over_m_vs_p}. As discussed in the previous subsections, for each possible bit-width $P_w$, we maximize the total parameter count $N$ such that the memory budget constraint in \cref{eq:mem_budget} is satisfied. We then plot $N_{\mathrm{eff}}(P_w) / M$.

\section{Theoretical model of software-decoded formats}
\label{sec:app-compute-model}

We wish to build a basic theoretical model of speedup, based on a roofline hardware model of a device that supports $\mathrm{R_{compute}}$ arithmetic operations (multiplies or adds) per second and can transfer $\mathrm{R_{transfer}}$ bytes per second from its memory.

The fundamental operation is the matrix multiplication $Y^T = W X^T$ where $Y$ and $X$ are a \texttt{bf16} matrices of shape $m \times h$, $m$ is a batch dimension, and $h$ is a hidden dimension. $W$ is a quantized parameter matrix of shape $h \times h$, with an average bits/parameter of $P_w$.

In many inference scenarios, we can assume that the activations $X$ and $Y$ can be retrieved from on-chip caches, so we need only consider the transfer of $W$ from memory. (We note that the results would not change substantially if this wasn't the case.) Therefore the time required for compute and parameter transfer are
\begin{align}
    t_{\text{compute}} &= 2 \cdot m \cdot h^2 / \,\mathrm{R_{compute}} \\
    t_{\text{transfer}} &= P_w \cdot h^2 / \,(8 \cdot \mathrm{R_{transfer}}),
\end{align}
where the factor of $2$ accounts for multiplies and adds separately, consistent with standard definitions of $\mathrm{R_{compute}}$ op/s, and the factor of $/8$ accounts for the measurement of $P_w$ in bits whereas $\mathrm{R_{transfer}}$ is commonly measured in bytes.

In our model, we assume ideal overlapping of communications and compute, in which case $t = \max(t_{\text{compute}}, t_{\text{transfer}})$. This gives
\begin{equation}
    t = \frac{2 \cdot m \cdot h^2}{\mathrm{R_{compute}}}
        \max\left(1,
            \frac{P_w \cdot \mathrm{R_{compute}}}{16 \cdot \mathrm{R_{transfer}}}
        \right).
\end{equation}
If we set $\nu \coloneq \mathrm{R_{compute}}/\mathrm{R_{transfer}}$, and take the ratio $t_1/t_2$ to find the speedup associated with switching from $P_{w1}$ to $P_{w2}$ \emph{with the same $\mathrm{R_{compute}}$}, we obtain \cref{eq:theoretical-speedup}.

\paragraph{Regimes}
As highlighted in \cref{fig:perf_theoretical}, we observe that \cref{eq:theoretical-speedup} predicts three performance regimes based on the batch size $m$.

\begin{enumerate}
    \item For small $m < \min(P_{w1}, P_{w2}) \cdot \nu / 16$, the $\text{speedup}_{1\rightarrow 2}$ is the ratio of the precisions $P_{w1}/P_{w2}$. Both settings are memory-bound, so the time taken is proportional to the model size in bytes.
    \item For intermediate $m$, the speedup scales $\propto 1/m$.
    \item For large $m > \max(P_{w1}, P_{w2}) \cdot \nu / 16$ the speedup is $1$ (no speedup), as both settings are compute-bound, requiring approximately the same amount of compute work.
\end{enumerate}

\section{Benchmarking}
\label{sec:app-benchmarking}

To demonstrate the feasibility of $4$-bit and $1$-bit nonlinear block-scaled weight formats, we evaluate the performance of optimized kernels for these formats, running on a contemporary inference-focused deep learning hardware accelerator. Our micro-benchmarks of a single kernel execution and whole-model benchmarks show strong speedups using $4$ and $1$-bit weights for a given model size at batch size $1$ (for single-user autoregressive generation without speculative sampling), and some speedup at batch size $16$, however no speedup by batch size $256$.

Aside from the input token embedding lookup, which is computationally trivial, all matrix parameters in the Llama-like transformer architecture are consumed by token-wise projections, i.e. matrix multiplications where the local batch size $m$ is the product of the sequence length (during context ``prefill'' only) and ``user'' batch size. Therefore the fundamental operation for low-precision inference in the transformer architecture is the fused dequantize-multiply, $Y^T = \text{dequantize}(\tilde{W}) X^T$, where $\tilde{W}$ is a tuple incorporating the quantized element data, block scales and nonlinear format centroids (if applicable).

We provide full \href{https://github.com/graphcore-research/fused-dequantisation-kernels}{kernel and benchmark code} for sake of reproducibility.

\subsection{Kernel implementations}

We explore multiple implementations of the fused dequantize-matrix-multiply operation. All are semantically similar to the following PyTorch code, which uses a $256$-element ($8$-bit index) lookup table to map each packed input byte to $2\!\times\!4$-bit values or $8\!\times\!1$-bit values, followed by block scaling to reproduce the original weights.

\Needspace{12\baselineskip}
\begin{minted}{py}
def dequant_matmul(x, Wq, Wscale, lut8):
    """
    x      :: batch_size x input_size x bf16
    Wq     :: output_size x (input_size//elements_per_byte) x uint8
    Wscale :: output_size x (input_size//block_size) x bf16
    lut8   :: 256 x elements_per_byte x bf16
    return :: batch_size x output_size x bf16
    """
    Wu = lut8[Wq.long()].flatten(start_dim=1)
    B = Wu.shape[1] // Wscale.shape[1]
    W = Wu.view(*Wscale.shape, B).mul(Wscale.unsqueeze(2)).flatten(start_dim=1)
    return x @ W.T
\end{minted}

A minor variation is to use \emph{deferred scaling}, where the block scaling is deferred until after the accumulation of unscaled weights across a single block:

\Needspace{5\baselineskip}
\begin{minted}{py}
def dequant_matmul_defer_scale(x, Wq, Wscale, lut8):
    Wu = lut8[Wq.long()].flatten(start_dim=1)
    B = Wu.shape[1] // Wscale.shape[1]
    z = torch.einsum("mGg,hGg->mhG", x.unflatten(1, (-1, B)), Wu.unflatten(1, (-1, B)))
    return torch.einsum("mhG,hG->mh", z, Wscale)
\end{minted}

Both methods produce similar results, with subtle differences since floating-point operations do not exactly follow distributivity/associativity. The original approach requires $2 \cdot m \cdot h^2 + h^2$ floating-point operations, while deferred scaling requires $2 \cdot m \cdot h^2 + 2 \cdot m \cdot h^2 / B$ floating-point operations. Which is better in practice depends on hardware and compiler support.

We evaluate three implementations based on this strategy:

\begin{itemize}
    \item Using \texttt{torch.compile(mode=max-autotune-no-cudagraphs)} on similar PyTorch code to the example shown above, with \texttt{bf16} compute.
    \item Custom Triton \citep{tillet2019triton} kernels for chunked matrix-vector and matrix-matrix. We rely on the Triton autotuner to select chunk shape hyperparameters from a set of useful configurations identified by a broad sweep on the evaluation GPU system. We also find empirically that in all preferred configurations it is beneficial to disable automatic pipelining with \texttt{num\_stages=1}, and to use \texttt{num\_warps=1} for most configurations. This method also uses \texttt{bf16} compute.
    \item An adaptation to the Marlin \citep{frantar2024marlinmixedprecisionautoregressiveparallel} $4$-bit linear matrix multiply with block scaling, where we add our $8$-bit LUT and adapt the code for $B=64$. This implementation only supports $4$-bit formats and uses \texttt{float16} for activations and compute. We note that the GPU system evaluated supports \texttt{float16} at the same speed as \texttt{bf16}.
\end{itemize}

\subsection{Micro-benchmarks}
\label{sec:app-benchmarking-micro}

For our micro-benchmarking evaluation, we test applicable implementations across a range of relevant $m$ (equivalent to batch size), $h$ (hidden dimension) and bit-widths $P_w$. We create a CUDA graph of $100$ calls to the kernel, using different input and output tensors for each call (to avoid unrealistic caching advantages). All calls take place on the same stream, so there is no overlap between executions. We invoke this CUDA graph $100$ times, reporting the average wall-clock time of a single kernel invocation and the standard error across runs.

Note that we report micro-benchmark results solely for square projections where $W$ is a $h \times h$ matrix. While some transformer model projections are commonly square (e.g. attention query and output projections), many are not (e.g. feedforward network up, gate and down projections). In early experiments, we observed that the key performance trends do not vary substantially for non-square projections, therefore we omit these results for brevity. Non-square projections are necessarily present in the model benchmarks of \cref{sec:app-benchmarking-model}.

\paragraph{Setup} The test platform is an NVIDIA L40S GPU, which has $864\gbs$ memory bandwidth and peak dense matrix compute of $362$ TFLOP/s in \texttt{bf16}. Applicable software versions are PyTorch 2.9.1, Triton 3.5.1 and CUDA 12.8.

\paragraph{Results} Headline results for batch size $m=1$ and a fixed problem size are shown in \cref{tab:perf_micro_m1_impl}. We expect good speedups in this case, since the kernel execution should be dominated by the time taken to transfer the weight matrix from memory. Effective bandwidth measures the operand sizes divided by the time taken, and can be used as an efficiency measure for such memory-bound kernels.

Both Triton and Marlin implementations are able to achieve good, almost optimal, speedups for $4$-bit dequantization using a lookup table and block scaling, but the $1$-bit Triton kernel is not able to saturate memory bandwidth in the same way, due to a smaller amount of data to transfer and more dequantization and compute work per byte transferred.

\begin{table}[h]
    \centering
    \caption{
    A micro-benchmark of different kernel implementations, for $m=1$ and $h=8192$. The theoretical maximum speedup for $4$-bit and $1$-bit kernels is $3.8\times$ and $12.8\times$ respectively. Standard errors are $< 0.2\, \mathrm{\upmu s}$ except for the torch.compile implementation ($< 0.6 \,\mathrm{\upmu s}$).
    }
    \label{tab:perf_micro_m1_impl}
    \begin{tabular}{llrr} \toprule
  $P_w$ & Implementation & Time (Speedup) & Effective Bandwidth \\\midrule
  $16$ & Torch & $181.6\,\mathrm{\upmu s}$ ($1.0\times$) & $739\,\mathrm{GB/s}$ \\
  $16$ & Triton & $175.6\,\mathrm{\upmu s}$ ($1.0\times$) & $764\,\mathrm{GB/s}$ \\
  $4.25$ & torch.compile & $772.3\,\mathrm{\upmu s}$ ($0.2\times$) & $46\,\mathrm{GB/s}$ \\
  $4.25$ & Marlin & $53.5\,\mathrm{\upmu s}$ ($3.4\times$) & $667\,\mathrm{GB/s}$ \\
  $4.25$ & Triton & $49.5\,\mathrm{\upmu s}$ ($3.7\times$) & $721\,\mathrm{GB/s}$ \\
  $1.25$ & torch.compile & $770.4\,\mathrm{\upmu s}$ ($0.2\times$) & $14\,\mathrm{GB/s}$ \\
  $1.25$ & Triton & $24.0\,\mathrm{\upmu s}$ ($7.6\times$) & $438\,\mathrm{GB/s}$ \\
\bottomrule
\end{tabular}
\end{table}

\begin{figure}[tp]
    \includegraphics[width=\textwidth]{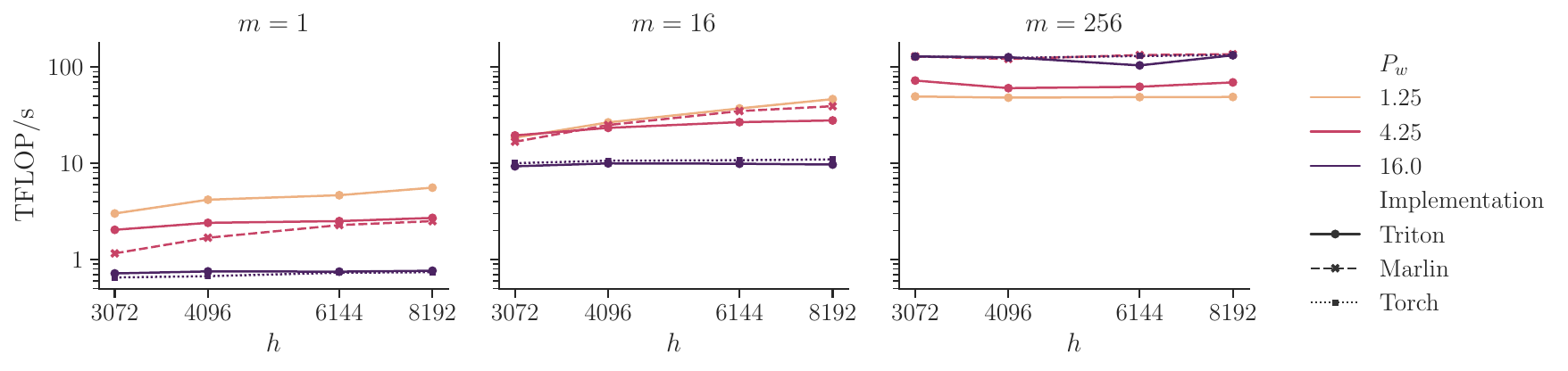}
    \caption{
    A micro-benchmark of kernel performance for square matrix multiplication, swept over batch size $m$, hidden size $h$ and bit-width. The maximum error bar ($\pm 2$ standard errors) is approximately $\pm 1\%$ of the mean, so error bars are not visible at this scale.
    }
    \label{fig:perf_micro_m_k_bits_impl}
\end{figure}

A broader sweep of results over various $m$, $h$ and bit-widths is shown in \cref{fig:perf_micro_m_k_bits_impl}. This shows that both Marlin and Triton implementations retain a (smaller) speedup over full \texttt{bf16} precision at batch size $m\!=\!16$, but the advantage is lost by $m\!=\!256$. This is consistent with the transition from the memory-bound small-batch regime to the compute-bound large-batch regime. The Marlin kernel is particularly strong at intermediate batch sizes. We note that a similar technique of custom CUDA combined with an optimized data layout would also be applicable to $1$-bit dequantization, opening the possibility for further improvements.

\subsection{Model benchmarks}
\label{sec:app-benchmarking-model}

We also evaluate the same kernel implementations in the context of autoregressive token generation, which is a memory-bound operation at small batch size due to the lack of sequence parallelism. We use the Llama 3 model implementation from transformers \citep{wolf2019huggingface}, with model configurations from \cref{sec:memory_matched_downstream}, which retain the embedding and final projection matrices in \texttt{bf16}, quantizing all other matrix parameters, and capture the forward pass in a CUDA graph. We generate $100$ tokens and record the average time per token. The hardware setup is as described in \cref{sec:app-benchmarking-micro}.

\paragraph{Results} Main results for batch size $1$ are shown in \cref{tab:perf_model_m1_impl}, showing that $4$ and $1$-bit formats give consistent speedups for a given model size, in this memory-bound setting. Importantly, the $4$-bit $12$B model is only $\approx 10\%$ slower than the $16$-bit $4$B model, however the $1$-bit $31$B model is somewhat slower. We also observe the small overhead of nonlinear quantization in Marlin by comparing against the original \emph{Marlin (Linear)} dequantization approach that uses arithmetic operations in place of a lookup table.

\begin{table}[h]
    \centering
    \caption{
    Model tokens/s for autoregressive generation at batch size $1$. Entries in {\color{blue}blue} correspond to our scaled-up training settings from \cref{sec:memory_matched_downstream} The $4$-bit $12$B model is $\approx 10\%$ slower than the $16$-bit $4$B model, while the $1$-bit $31$B model has somewhat lower tokens/s since the dequantization and higher-precision arithmetic is unable to saturate memory bandwidth (see \cref{tab:perf_micro_m1_impl}). The maximum error bar ($\pm 2$ standard errors) is $<\!1\%$ of the mean.
    }
    \label{tab:perf_model_m1_impl}
    \begin{tabular}{llrrr} \toprule
  $P_w$ & Implementation & 4B & 12B & 31B \\\midrule
  $16$ & Torch & \color{blue}{$88.8$} & $30.0$ & OOM \\
  $4.25$ & Triton & $190.1$ & \color{blue}{$79.3$} & $35.4$ \\
  $4.25$ & Marlin & $151.5$ & $64.1$ & $32.4$ \\
  $4.25$ & Marlin (Linear) & $156.2$ & $65.5$ & $32.8$ \\
  $1.25$ & Triton & $245.6$ & $120.0$ & \color{blue}{$60.7$} \\
\bottomrule
\end{tabular}
\end{table}

\begin{figure}[tp]
    \includegraphics[width=\textwidth]{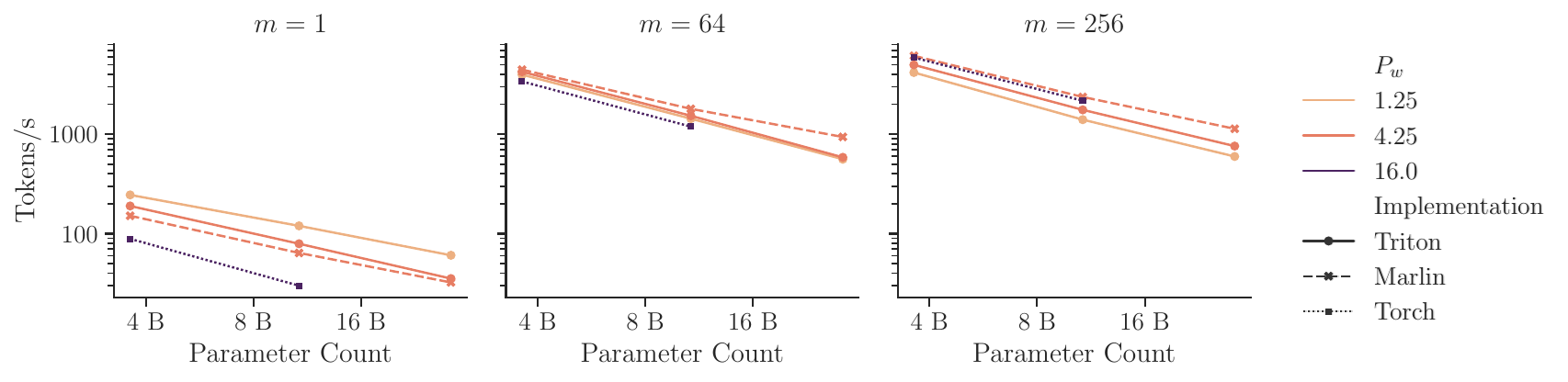}
    \caption{
    Autoregressive decoding speed for different model size, batch size and bit-width. Larger batches are beneficial to token/s, since they amortize the cost of transferring the model parameters, but this has greater impact on the \texttt{bf16} baseline than on compact formats. The maximum error bar ($\pm 2$ standard errors) is $<\!0.5\%$ of the mean.
    }
    \label{fig:perf_model_params_m_bits_impl}
\end{figure}

Further results are shown in \cref{fig:perf_model_params_m_bits_impl}. Consistent with micro-benchmarking outcomes, the advantage of weight quantization reduces at batch size $64$ and disappears by batch size $256$. This shows the applicable regime for formats without hardware support, which can accelerate execution during small batch token generation, but can instead impose some overhead at large batch sizes, including context prefill. In such settings, it may be more efficient to dequantize in a separate kernel, before using a standard high-precision GEMM kernel for the compute.

\end{document}